\documentclass{article}
\usepackage[preprint]{colm2026_conference}

\usepackage[T1]{fontenc}
\usepackage[utf8]{inputenc}
\usepackage{microtype}
\usepackage{hyperref}
\usepackage{url}
\usepackage{booktabs}
\usepackage{amsfonts}
\usepackage{nicefrac}
\usepackage{lineno}
\usepackage{graphicx}
\usepackage{float}
\usepackage{multirow}
\usepackage{placeins}
\usepackage{enumitem}
\usepackage{arydshln}
\usepackage{tcolorbox}
\usepackage{verbatim}
\usepackage{listings}
\usepackage{xcolor}
\usepackage{caption}
\usepackage{wrapfig}
\usepackage{amsmath}

\definecolor{darkblue}{rgb}{0, 0, 0.5}
\hypersetup{colorlinks=true, citecolor=darkblue, linkcolor=darkblue, urlcolor=darkblue}

\newcommand{\ie}{\emph{i.e.}, }
\newcommand{\eg}{\emph{e.g.}, }

\newcommand{\change}[1]{\textcolor{black}{{#1}}}

\newtcolorbox{prompt}[1]{
    colback=gray!20,
    colframe=black,
    boxrule=0.3pt,
    arc=3mm,
    left=2pt,
    right=2pt,
    boxsep=3pt,
    fonttitle=\small\bfseries,
    title=#1,
    fontupper=\scriptsize
}

\newtcolorbox{prompt_two_column}[1]{
    colback=gray!20,
    colframe=black,
    boxrule=0.3pt,
    arc=3mm,
    left=2pt,
    right=2pt,
    boxsep=3pt,
    fonttitle=\small\bfseries,
    title=#1,
    fontupper=\scriptsize,
    floatplacement=t,
    float*=true
}

\title{Idea2Plan: Exploring AI-Powered Research Planning}

\author{
Jin Huang$^{1}\thanks{Work performed while at Microsoft Research.}$ \quad
Silviu Cucerzan$^{2}$ \quad
Sujay Kumar Jauhar$^{2}$ \quad
Ryen W. White$^{2}$ \\
$^{1}$University of Michigan \quad
$^{2}$Microsoft Research
}

\begin{document}

\ifcolmsubmission
\linenumbers
\fi

\maketitle

\begin{abstract}
Large language models (LLMs) have demonstrated significant potential to accelerate scientific discovery as valuable tools for analyzing data, generating hypotheses, and supporting innovative approaches in various scientific fields.
In this work, we investigate how LLMs can handle the transition from conceptual research ideas to well-structured research plans. Effective research planning not only supports scientists in advancing their research but also represents a crucial capability for the development of autonomous research agents. Despite its importance, the field lacks a systematic understanding of LLMs' research planning capability.
To rigorously measure this capability, we introduce the \emph{Idea2Plan} task and \emph{Idea2Plan Bench}, a \change{set of benchmarks} built from ICML 2025 \change{and Nature Mental Health papers} released after major LLM training cutoffs. Each benchmark instance includes a research idea and a grading rubric capturing the key components of valid plans.
We further propose \emph{Idea2Plan JudgeEval}, a complementary benchmark to assess the reliability of LLM-based judges against expert annotations.
Experimental results show that GPT-5 achieves the strongest performance on the benchmark, though substantial headroom remains for improvement. Our study provides new insights into LLMs' capability for research planning and lays the groundwork for future progress.\footnote{An independently created dataset based on this paper is available at \url{https://github.com/Jn-Huang/idea2plan_icml2026}.}
\end{abstract}

\section{Introduction}

A key challenge in scientific research is that scientists tend to produce more promising ideas than they can pursue.
When attending conferences, participating in reading groups, or discussing with peers, scientists frequently conceive ways to improve ongoing work or apply insights to other areas. Despite this abundance of ideas, many remain unexplored due to time constraints.

Transforming each research idea into a viable plan requires considerable time and cognitive effort. The process of developing an idea often involves reviewing prior work, formulating hypotheses, selecting appropriate methods, and designing experiments, with plans iteratively refined as research progresses.
This process can take researchers many days or even weeks to complete.
We refer to the process of turning a research idea into a concrete, testable plan as \emph{research planning}. Research planning encompasses all the steps necessary to bridge the gap between an initial idea and a well-structured plan ready for execution.

Automated systems capable of research planning could greatly accelerate scientists' progress. By supporting the development of research ideas and streamlining the planning process, such systems could also advance the capabilities of autonomous AI research agents and help to ensure that more promising ideas are explored and developed.

Despite its importance, AI-powered research planning has been underexplored. Recent works focus on other research stages such as idea generation~\citep{SciMON,SciAgents,ResearchAgent,llm_novel_idea}, literature review~\citep{auto_survey}, and experiment execution~\citep{PaperBench,Curie,CodeScientist,paper2code}. Previous work treats research planning as an intermediate step without explicit evaluation~\citep{CodeScientist} or evaluates research plans based only on simple traits such as clarity and novelty~\citep{MLR-Bench}. 

\begin{figure}[t]
   \centering
   \includegraphics[width=\textwidth]{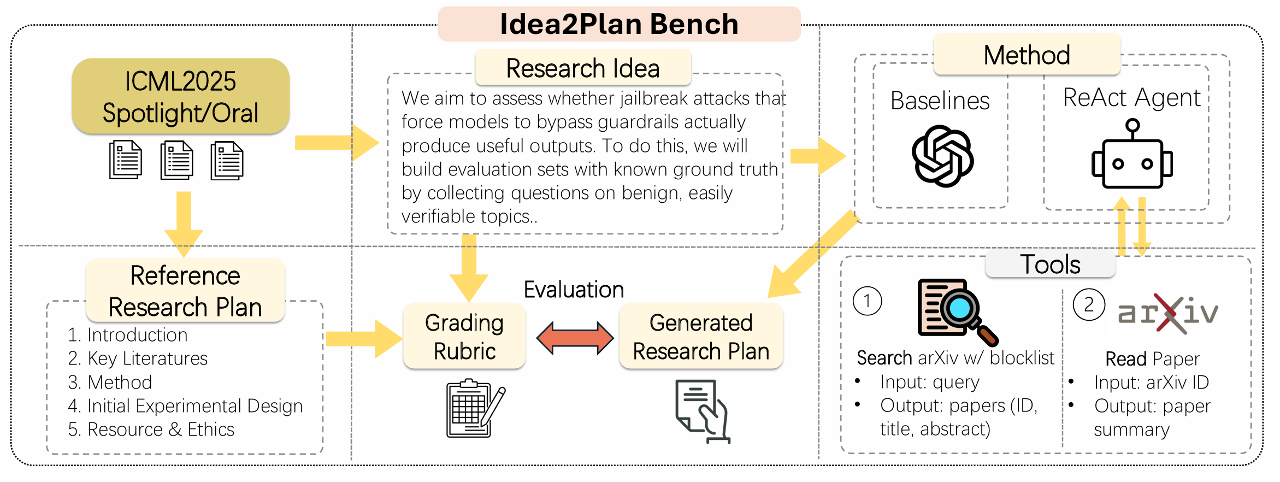}
   \vspace{-6mm}
   \caption{Idea2Plan overview: Starting from a research idea extracted from the abstract of an academic paper, we employ a pipeline that includes prompt-based and ReAct agent methods, integrated with search and paper reading tools, to generate research plans that are then evaluated against reference plans derived from the content of the corresponding paper by using a comprehensive grading rubric.}
   \label{fig:idea2plan_bench}
\end{figure}

We make the evaluation of AI’s ability for research planning our primary objective.
We start our investigation by formulating the \emph{Idea2Plan} task, where the input is a research idea, and the AI agent is asked to generate a \emph{research plan}, a structured output that serves as a roadmap for executing a project. 
A research plan typically includes objectives, prior work, methodology, and evaluation design~\citep{weber2008developing,Sudheesh2016}.
\change{For AI research specifically,} we instantiate our research plans' structure based on common AI paper conventions: Introduction, Key Literature, Methods, Initial Experimental Design, and Resources/Compliance/Ethical Considerations. We hypothesize that this format mirrors how AI researchers articulate and reason about research plans.

Evaluating such plans is challenging for two major reasons. 
First, data contamination threatens validity, because an LLM may have already encountered the idea (for instance through a published paper), in which case its output reflects the LLM's memorization about the research idea rather than genuine research planning ability~\citep{carlini2021extracting, carlini2022quantifying}. To mitigate this, we build an evaluation framework that extracts research ideas from AI papers that are published \emph{after} LLMs' data cutoff dates. Our design allows re-running the same pipeline after future model updates, producing new evaluation datasets that remain contamination-free.  
Second, multiple plans can be equally valid for the same idea; for example, two sound plans may select different yet comparable datasets, yielding distinct but acceptable designs.  
We address this by constructing a dedicated grading rubric for each research idea. The questions in the rubric test the common desiderata that any valid plan for that idea should meet.

Concretely, to instantiate the \emph{Idea2Plan} task for empirical evaluation, we construct \emph{Idea2Plan Bench} based on ICML 2025 Spotlight and Oral papers, as illustrated in Figure~\ref{fig:idea2plan_bench}.
We exclude papers with public arXiv versions before the latest training data cutoff of the LLMs employed in our study.
We then generate a grading rubric from the extracted research plan for each research idea. 
The rubric contains five sections of yes/no questions that capture the essential components of a valid plan for each research idea as derived from the content of the corresponding paper.
We assess the quality of extracted plans and rubrics by a human study and find that they are of reasonable quality.
To test the generalizability of our pipeline beyond AI research, we also create a benchmark from Nature Mental Health papers using the same pipeline.

Given the amount of effort required to grade research plans against rubrics, it is unrealistic to rely on human graders at scale. To make evaluation scalable, we propose LLM-based judges for grading the rubrics~\citep{llm_judge, PaperBench}. We introduce an additional benchmark, \emph{Idea2Plan JudgeEval}, to assess judgment reliability. This benchmark collects a set of ground-truth gradings created manually by human experts.
Our results show that LLM judges can provide expert-aligned grading.

To explore the performance of LLMs on the \emph{Idea2Plan} benchmark, we test both direct prompting and agentic approaches. Our agentic approach uses a ReAct-style scaffolding~\citep{ReAct} with tools for searching over and reading arXiv papers. We then compare a range of frontier proprietary and open-source LLMs. Results show that GPT-5 and GPT-5-mini consistently outperform other LLMs. Surprisingly, the ReAct agent's performance does not outperform direct prompting.
Overall, our work advances the understanding of research planning capabilities in LLMs.

Our contributions are as follows:
\begin{itemize}[nosep]
\item We formalize the \emph{Idea2Plan} task and \change{develop a pipeline to automatically generate rubrics for evaluating research plans. This pipeline allows us to produce new contamination-free benchmarks after future model updates.}

\item We build a \change{set of benchmarks} from ICML 2025 \change{and Nature Mental Health papers}. Human study shows that the extracted plans and rubrics are of \change{reasonable} quality.

\item We propose an LLM-based judge and the \emph{Idea2Plan JudgeEval} to compare LLM-as-judge scoring with expert annotations.
\item We discuss experiments and results with several state-of-the-art LLMs under different prompting and agentic strategies. 
\end{itemize} 

\section{Related Work}

\subsection{AI Agents for Scientific Discovery}
AI agents increasingly support different stages of the research workflow, including hypothesis generation~\citep{SciMON, SciAgents, ResearchAgent, llm_novel_idea, MASSW, CodeScientist, co-scientist}, literature review~\citep{ai2_scholar_qa, auto_survey}, and experimental execution~\citep{Curie, CodeScientist, paper2code}. 

Recent efforts further pursue end-to-end AI scientists that can conduct a full loop of the scientific process from generating ideas to autonomously designing and conducting experiments~\citep{dolphin,ai-scientist,ai-scientist-v2,agent-lab,AgentRxiv}. For example, the \textsc{AI Scientist} proposes a pipeline to generate a research idea, conduct experiments, and finally write a paper~\citep{ai-scientist}.
Despite the advances of these systems, most prior research does not treat planning—a phase that bridges ideation and implementation—as a first-class objective. Research planning is crucial, as it directly affects downstream execution quality. We therefore focus our study on evaluating this critical yet underexplored capability.

\subsection{Benchmarks for AI Agents on Research Automation}

Recent benchmarks aim to evaluate AI systems designed for research automation. Broadly, these efforts can be grouped into two categories based on their \textit{output form}: (1) paper-level evaluation, where systems are assessed on the quality of generated research papers~\citep{ai-scientist, ai-scientist-v2, agent-lab}; and (2) code-level evaluation, where systems are judged by their ability to produce code implementations or experimental results~\citep{Curie, ScienceAgentBench, CodeScientist, PaperBench, paper2code,MLR-Bench, EXP-Bench}. 
In contrast, our work isolates and rigorously evaluates the stage of research planning. The most related works are \textsc{CodeScientist}~\citep{CodeScientist} and \textsc{MLR-Bench}~\citep{MLR-Bench}: the former includes research planning as an intermediate step without explicit evaluation, and the latter evaluates research plans only on high-level dimensions such as clarity and novelty.

\subsection{Planning}

Planning, which refers to creating a series of actions to achieve an objective, is an important capability for both humans and AI agents~\citep{newell1958elements, automated_planning}. Conventional studies primarily depend on symbolic approaches or reinforcement learning methods~\citep{pddl, intro_planning, he2015deep}. Recently, there has been increasing interest in using LLMs in planning~\citep{llm_planning_survey, llm_agent_planning_survey}.
Recent approaches demonstrate that LLMs can explicitly decompose goals into reasoning steps or structured action sequences, enabling them to handle long-horizon, multi-step tasks~\citep{ReAct, Plan-and-Solve, generalized_planning_in_pddl}.
Despite these advances, LLMs' ability to perform research planning remains largely unexplored, which we seek to address in this paper.

\section{Proposed Framework}

\begin{table}[!t]
\centering
\caption{We show an example of a research idea, grading questions, relevant excerpts of a research plan generated by GPT-5 given this research idea, and the LLM-judge's judgements of this question. The full example is in \S\ref{appen:jailbreak_tax_example}.}\vspace{-1mm}
\label{tab:main_paper_example_rubric_etc}
\small
\begin{tabular}{|p{0.95\textwidth}|}
\hline
\textbf{Research Idea (from paper: The Jailbreak Tax: How Useful are Your Jailbreak Outputs?~\citep{jailbreak_tax}, ICML 2025 Spotlight)} \\
\hline
``We aim to assess whether jailbreak attacks that force models to bypass guardrails actually produce useful outputs. To do this, we will build evaluation sets with known ground truth by collecting questions on benign, easily verifiable topics and align models to refuse those questions. We will then apply representative jailbreak strategies to these aligned models and measure how much model utility drops in the jailbroken responses. We propose a new metric to quantify this performance degradation and introduce corresponding benchmarks to enable systematic evaluation and comparison of jailbreak methods.'' \\
\hline
\end{tabular}
\vspace{0.3cm}
\scriptsize
\setlength{\tabcolsep}{3pt}
\begin{tabular}{|p{1.2cm}|p{5.0cm}|p{5.5cm}|p{0.8cm}|}
\hline
\textbf{Section} & \textbf{Grading Question Example} & \textbf{Excerpts from a Research Plan generated by GPT-5} & \textbf{Judg-ment} \\
\hline
Introduction &
``Does the plan note that existing evaluations focus on bypass success and neglect post-jailbreak capability retention?'' &
``While jailbreak success is typically measured by refusal circumvention, it is unclear whether the resulting outputs remain useful and accurate.'' &
Yes \\
\hline
Key\newline Literatures &
``Does the plan cite the paper (Jailbreaking black box large language models in twenty queries) \underline{or similar work} on iterative LLM-based prompt rewriting attacks?'' &
``Greshake et al., Prompt Injection attacks [...] Wallace et al., Universal Adversarial Triggers [...] Zou et al., adversarial suffix (GCG) methods.'' &
No \\
\hline
Methods &
``Does the plan define a metric for post-jailbreak utility, such as conditional task accuracy after a successful jailbreak?'' &
``Conditional Answer Correctness (CAC): accuracy among non-refusal responses under attack.'' &
Yes \\
\hline
Initial\newline Exp.\newline Design &
``Does the plan include analysis of the relationship between task difficulty and the magnitude of the jailbreak tax?'' &
``Optional: chain-of-thought vs concise answers to see if jailbreaks disproportionately harm multi-step reasoning.'' &
No \\
\hline
Resources,\newline Compl.,\newline \& Ethics &
``Does the plan address the potential for adversaries to misuse the evaluation framework?'' &
``Release attack code in a restricted form that limits adaptation to harmful content [...] exclude especially potent per-item adversarial suffixes from public artifacts.'' &
Yes \\
\hline
\end{tabular}
\vspace{-4mm}
\end{table}

In this section, we introduce \emph{Idea2Plan}, a framework to evaluate how LLMs turn research ideas into research plans; see Figure~\ref{fig:idea2plan_bench} for an overview.

\subsection{Research Plan and Idea Extraction}
\label{sec:research_plan_template}

There may be different definitions of research plans and what should be included~\citep{weber2008developing,Sudheesh2016}. In this work, we focus on AI research and we adhere to the common structure of research plans for AI papers. We discard plan components that are not easily determined by an LLM (\eg timelines, logistics, or human expertise). We guide the research plan generation using a structured template with the following sections: Introduction, Key Literature, Methods, Initial Experimental Design, and Resources/Compliance/Ethical Considerations. We use o4-mini\footnote{\url{https://openai.com/index/introducing-o3-and-o4-mini/}} (reasoning=high) for idea and plan extraction. Prompt templates are shown in Appendix \S\ref{appen:dataset_plan_idea}.

\subsection{Rubric-based Evaluation}

Given our extracted reference research plans and research ideas, along with research plans generated by LLMs (either through prompting or agentic approaches), the question is: how do we evaluate the generated research plans? The key difficulty lies in the fact that a single research idea can lead to multiple equally reasonable research plans. For example, an idea about evaluating a new jailbreak attack could reasonably choose different yet comparable datasets, threat models (\emph{e.g.,} black-box vs. white-box), or metric suites; these choices can produce distinct plans that are all sound.

To capture these nuances, we propose a rubric-based evaluation approach~\citep{PaperBench}. We generate rubric questions from the reference research plans to identify high-level design choices that should be addressed in any reasonable research plan for the given idea. This approach allows us to evaluate whether generated plans cover essential conceptual elements while accommodating legitimate variations. The prompt template is in \S\ref{appen:dataset_rubric}.

\subsection{Dataset Construction}

Our dataset selection follows two key criteria: (1) papers must be free from potential training data contamination for the LLMs we evaluate, and (2) the research ideas must be of high quality.
To meet these requirements, we propose to use the Spotlight and Oral papers from ICML 2025, a top-tier AI conference. We filter out any papers with arXiv submissions predating the most recent data cutoff of the LLMs we test (GPT-5, October 2024\footnote{\url{https://platform.openai.com/docs/models/gpt-5}}).
From this filtered set, we randomly select 200 papers as our test set. Additionally, we randomly sample 30 papers as a development set, keeping the test set reserved exclusively for final evaluation.
We also construct a dataset for a different scientific domain from Nature Mental Health\footnote{\url{https://www.nature.com/natmentalhealth/}} papers using the same pipeline (Appendix~\S\ref{appen:nmh_dataset}).

\begin{table}[t]
\centering
\begin{minipage}[t]{0.42\textwidth}
\centering
\scriptsize
\caption{Expert evaluation of the extracted research plans and rubrics, rated on a five-point Likert scale (1 = major issues, 5 = well done overall).}
\label{tab:expert_evaluation_scores}
\begin{tabular}{lcc}
\toprule
\textbf{Criterion} & \textbf{Plan} & \textbf{Rubrics} \\
\midrule
Introduction & $4.12 \pm 0.83$ & $4.00 \pm 1.07$ \\
Related Lit. & $4.00 \pm 0.76$ & $4.12 \pm 0.99$ \\
Method & $4.06 \pm 1.08$ & $4.00 \pm 0.93$ \\
Experiments & $4.25 \pm 0.71$ & $4.50 \pm 0.76$ \\
Resource & $4.50 \pm 0.76$ & $4.62 \pm 0.52$ \\
Ethics & $4.88 \pm 0.35$ & $4.75 \pm 0.46$ \\
\bottomrule
\end{tabular}
\end{minipage}%
\hfill
\begin{minipage}[t]{0.56\textwidth}
\centering
\scriptsize
\caption{Performance of LLM judges against human annotations on 30 plans. Metrics are macro-averaged across papers. \emph{Cost} denotes average API cost (\$) to grade one plan.}
\label{tab:judgeeval}
\begin{tabular}{lccccc}
\toprule
\textbf{Judge Model} & \textbf{Acc.} & \textbf{Prec.} & \textbf{Rec.} & \textbf{F1} & \textbf{Cost} \\
\midrule
GPT-4.1-mini & .830 & .771 & .878 & .821 & \$0.01 \\
GPT-4.1 & .822 & .744 & .911 & .819 & \$0.07 \\
o4-mini & .896 & .858 & .918 & .887 & \$0.10 \\
GPT-5-mini & .833 & .736 & \textbf{.970} & .837 & \$0.07 \\
GPT-5 & \textbf{.943} & \textbf{.918} & .956 & \textbf{.937} & \$0.17 \\
\bottomrule
\end{tabular}
\end{minipage}
\end{table}

\subsection{Expert Assessment of Extracted Research Plans and Rubrics}
\label{sec:expert_assessment}

To validate the quality of our extraction pipeline, we conduct a human evaluation study with eight Ph.D. students and researchers with expertise in AI. Each expert is asked to select one paper from their area. This covers domains including computer vision, natural language processing, graph machine learning, AI for science, and theoretical machine learning. The assessment consists of two components:
\begin{enumerate}
     \item Experts are presented with the LLM-extracted research plan alongside the original paper, with instructions such as: “\emph{Please rate how accurately the LLM captured the content from the original paper.}” Ratings are provided on a five-point Likert scale.
    \item Experts are shown the extracted rubric questions and the corresponding research idea, then asked to ``\emph{assess whether the questions cover the essential parts of a research plan for the given idea}'' on a five-point Likert scale.
\end{enumerate}

We present the expert evaluations in Table~\ref{tab:expert_evaluation_scores}. Across sections, mean scores exceed 4.0 out of 5 (scale anchors: 1 = major issues, 2 = significant problems, 3 = average, 4 = acceptable with some issues, 5 = well done overall). 
These scores suggest that the extracted research plans and rubrics are of reasonable quality (more details in \S\ref{appen:expert_papers}). 

\subsection{Grading and LLM-based Judges }

The rubric for each paper consists of five sections and there is a list of yes/no questions to specify the criteria that a research plan should satisfy. After grading each question in a section, we compute binary classification accuracy—the proportion of rubric questions judged as satisfied. This accuracy represents the \emph{Planning Score} for that section. We then take the macro-average of the five section accuracies as the final \emph{Planning Score} for this paper.
Finally, we compute the overall score by averaging these paper-level scores across all papers, which corresponds to a macro-average over papers. Our main metric is therefore the \textbf{Average Planning Score} across all papers.

Since manually grading each generated research plan against our rubrics would be prohibitively expensive, we employ LLM-based judges~\citep{llm_judge, PaperBench}. Specifically, we prompt an LLM with the generated research plan and the rubric, and instruct the LLM to generate a grading (\ie give a yes/no judgment for each rubric question). We discuss how we assess the quality of LLM-based judges in the next section.
We present the prompt templates in \S\ref{appen:llm_judge}. 

\subsection{Evaluating LLM Judges with JudgeEval}

We collect a dataset (\emph{Idea2Plan JudgeEval}) to assess the performance of different LLM judges against human expert annotations. \change{We randomly sample 30 papers from Idea2Plan Bench and, for each paper, randomly select one generated research plan from our experiments (covering diverse LLM and baseline combinations). Two annotators independently grade all rubric questions for each plan, achieving 88.4\% inter-annotator agreement (Cohen's $\kappa = 0.768$). For disagreements, a third annotator makes the final judgment.} Since each rubric question is a binary classification task (a yes/no question), we report standard binary classification metrics with macro-averaging to aggregate performance across papers.

\change{The results show that GPT-5 achieves the highest F1 score (0.937), while GPT-4.1-mini offers the most cost-effective option. All models exhibit a leniency bias, producing more false positives than false negatives.} We select o4-mini (reasoning=high) as our evaluation model.

\subsection{Baseline Choices and Agent Design}

We evaluate several baselines for research plan generation, ranging from simple prompting to agentic frameworks with external tools.

\paragraph{Naïve Baseline.} Our simplest baseline uses a minimal prompt (\textit{``Your task is to generate a detailed research plan based on the provided research idea.''}) along with section titles in the research plan template to constrain the generation format. 
This baseline mimics the most straightforward scenario where scientists directly prompt an LLM for research plans without any additional instructions.

\begin{figure}[t]
    \centering
    \includegraphics[width=\textwidth]{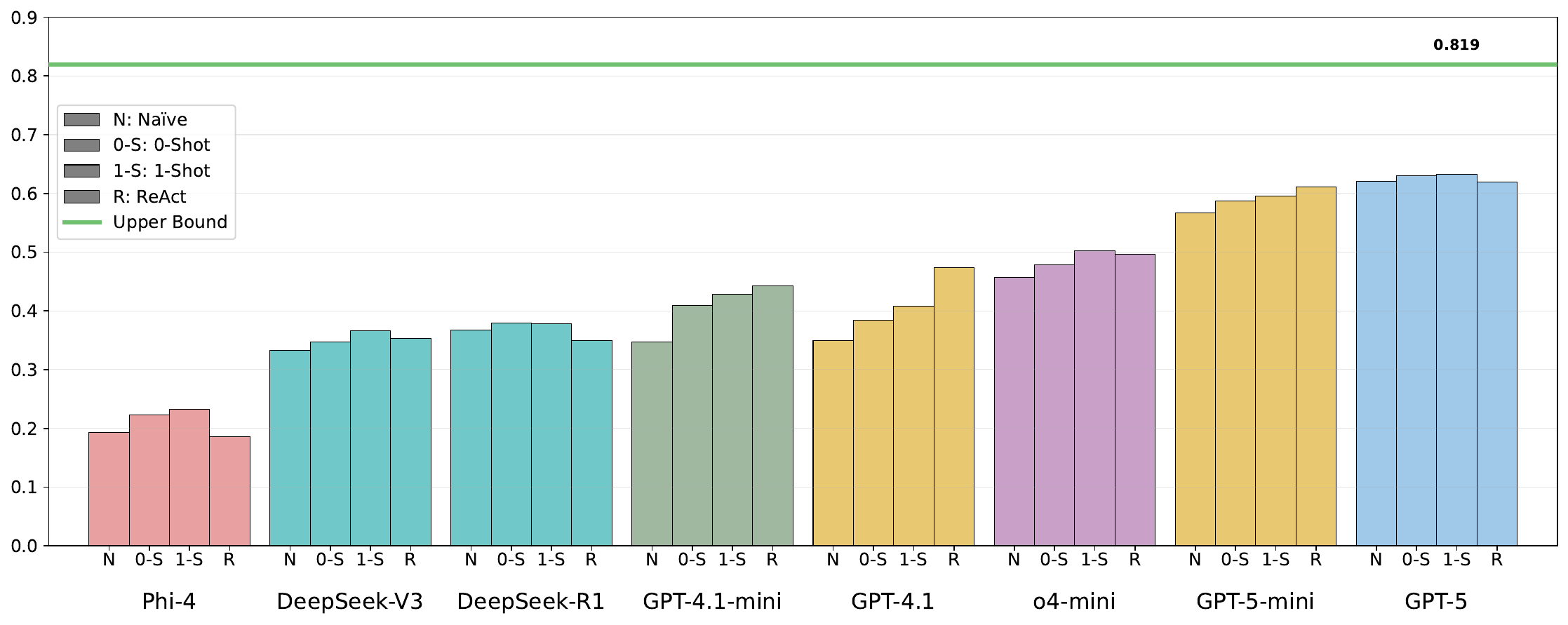}
    \caption{Average Planning Score for LLMs under four baselines (Naïve, 0-Shot, 1-Shot, ReAct). Scores are means over three independent runs per paper. We also include an \emph{upper-bound} in which o4-mini is given the original paper to generate a plan, approximating the highest achievable performance on \emph{Idea2Plan Bench}.}
    \label{fig:idea2plan_model_comparison}
\end{figure}

\begin{table}[t]
\centering
\scriptsize
\caption{Section-wise Planning Scores (\%) for Naïve and ReAct baselines. Each value represents the mean accuracy across all papers. Results for 0-shot and 1-shot baselines are provided in Table~\ref{tab:mean_scores_full}.}
\label{tab:naive_react_side}
\begin{tabular}{l:rrrrr:r:rrrrr:r}
\toprule
\multirow{2}{*}{Model} & \multicolumn{6}{c:}{Naïve} & \multicolumn{6}{c}{ReAct} \\
\cmidrule(lr){2-7} \cmidrule(lr){8-13}
 & Intro & Lit & Met & Exp & Res/Eth & Avg & Intro & Lit & Met & Exp & Res/Eth & Avg \\
\midrule
Phi-4        & 29.0 & 5.6 & 20.2 & 10.9 & 30.9 & 19.3 & 24.2 & 11.8 & 15.5 & 9.9  & 31.7 & 18.6 \\
DeepSeek-V3  & 39.8 & 25.2 & 32.4 & 24.7 & 44.5 & 33.3 & 39.9 & 23.8 & 33.4 & 28.5 & 50.9 & 35.3 \\
DeepSeek-R1  & 40.6 & 27.1 & 37.0 & 28.7 & 50.3 & 36.7 & 38.7 & 23.1 & 33.2 & 27.9 & 52.0 & 35.0 \\
GPT-4.1-mini & 42.3 & 21.8 & 34.7 & 27.4 & 47.2 & 34.7 & 50.6 & 30.6 & 44.5 & 35.7 & 60.3 & 44.3 \\
GPT-4.1      & 40.7 & 22.4 & 34.2 & 28.9 & 48.7 & 34.9 & 50.6 & 35.6 & 44.9 & 40.5 & 65.6 & 47.4 \\
o4-mini      & 50.8 & 29.1 & 50.2 & 39.1 & 59.7 & 45.7 & 54.2 & 36.1 & 54.4 & 42.6 & 61.2 & 49.6 \\
GPT-5-mini   & 58.5 & 38.1 & 65.9 & 51.6 & 69.5 & 56.7 & \textbf{62.3} & 44.5 & 67.3 & 55.2 & \textbf{76.4} & 61.1 \\
GPT-5        & \textbf{60.9} & \textbf{47.7} & \textbf{70.9} & \textbf{56.7} & \textbf{73.8} & \textbf{62.0} & 61.4 & \textbf{48.1} & \textbf{68.4} & \textbf{55.7} & 75.9 & \textbf{61.9} \\
\bottomrule
\end{tabular}
\end{table}

\paragraph{0-shot and 1-shot Baseline.} We enhance the basic prompt with additional instructions. For the 0-shot, we add general guidance such as \textit{``Generate a complete research plan following the EXACT template structure above.''} The 1-shot additionally includes one fixed research plan example to demonstrate the desired format and content structure.

\paragraph{ReAct Agent Baseline.} When developing research ideas, human scientists often review related literature and refine their plans by learning how others have addressed similar problems. Inspired by this, we evaluate an agentic framework that can access external information. We use a simple ReAct scaffolding following the ``Thought → Action → Observation'' loop~\citep{ReAct}. To simulate an environment where scientists interact with the research literature, we introduce two tools integrated with arXiv:
\begin{itemize}[nosep]
\item \textbf{ArXiv Search Tool with Blocklists.} Queries arXiv for relevant papers while preventing data contamination by applying two layers of blocklisting for each target paper: (1) exclude all papers published after January 30, 2025 (ICML 2025 deadline), and (2) exclude the target paper and all papers that cite it. The blocklisting is necessary because our small-scale experiments show that search engines can retrieve the target paper when given its research idea as a query. The tool takes an agent-generated query as input and returns a list of relevant arXiv papers with their titles, abstracts, and arXiv identifiers. We implement it using Bing Search\footnote{\url{https://learn.microsoft.com/en-us/azure/ai-foundry/agents/how-to/tools/bing-grounding}} restricted to the arXiv domain.
\item \textbf{ArXiv Read Tool.} Takes an arXiv identifier as input and returns a summary of the corresponding paper. We fetch the paper using the arXiv API and then use o4-mini to generate its summary with a fixed template that contains main contributions, key related literature, methods and techniques, experimental design and results. In this way, we avoid using the full content of a paper, as it would consume too much context and, as shown in our small-scale experiments, may reduce performance.
\end{itemize}
We limit the agent to at most five searches and five reads. The agent is free to decide the order of tool calls and whether to use all of the available calls. Prompt templates and design details are in \S\ref{appen:all_prompt_baselines_react}.

\section{Experiments}

\subsection{Experimental setup}

We test extensively on proprietary LLMs (GPT-4.1 and GPT-4.1-mini,\footnote{\url{https://openai.com/index/gpt-4-1/}} 
o4-mini,\footnote{\url{https://openai.com/index/introducing-o3-and-o4-mini/}} GPT-5 and GPT-5-mini,\footnote{\url{https://openai.com/gpt-5/}}) and open-source LLMs (DeepSeek-V3~\citep{deepseek-v3}, DeepSeek-R1~\citep{deepseek-r1} and Phi-4~\citep{phi-4}). For each model and baseline configuration, we run three independent trials per idea and report the mean performance.
We also include an \textit{upper-bound}, where o4-mini is prompted with the original paper and asked to generate a research plan. This serves as an estimate of the highest achievable performance on \emph{Idea2Plan Bench}.

\subsection{Results Analysis}

We present the average scores across all models and settings in Figure~\ref{fig:idea2plan_model_comparison} and the section-wise results in Table~\ref{tab:naive_react_side}. The full results are in \S\ref{appen:full_results}. We summarize the key findings below.

\paragraph{Naïve baseline achieves strong performance.} Despite being simple, the naïve baseline (one line prompt with section titles) attains performance close to the best methods across different models. The 0-shot and 1-shot baselines consistently outperform the naïve baseline, demonstrating that clear instruction and example provide meaningful improvements.

\paragraph{GPT-5 and GPT-5-mini lead the performance.} We notice that GPT-5 and GPT-5-mini substantially outperform other models. To understand this better, we compute pairwise win rates, \ie for each pair of models, the fraction of papers in which one model’s research plan scores higher than the other.
We find that GPT-5 and GPT-5-mini achieve over 90\% win rates against other models, indicating consistent superiority (see Figure~\ref{fig:win_rate_matrices}). 

\paragraph{Literature review is the most difficult section, while method and experiment sections show large differences across models.}  
As shown in Table~\ref{tab:naive_react_side}, Planning Scores for the literature section are the lowest across LLMs, reflecting the challenge of accurately identifying relevant prior work. The method and experiment sections exhibit the largest differences across LLMs, serving as clearer indicators of planning ability. GPT-5 and GPT-5-mini perform strongly across sections.

\begin{figure}[t]
    \centering
    \begin{minipage}[t]{0.62\textwidth}
        \centering
        \includegraphics[width=\textwidth]{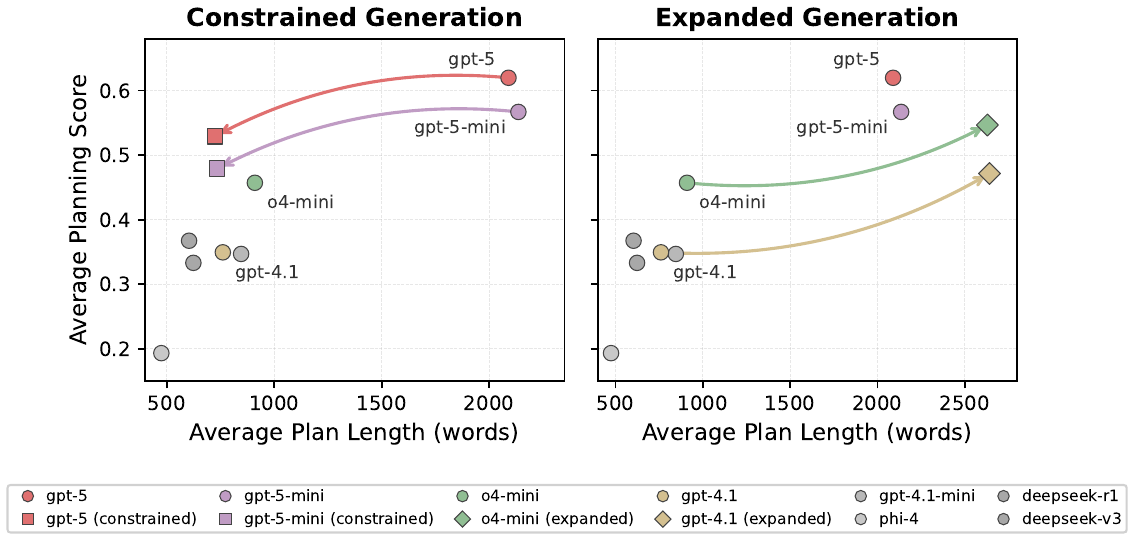}
        \caption{Relationship between plan length and Average Planning Score under constrained and expanded generation settings. \textbf{Left:} When response length is controlled, GPT-5 and GPT-5-mini generate shorter plans but remain the top performers. \textbf{Right:} When longer outputs are allowed, GPT-4.1 and o4-mini produce more extensive plans with higher scores. Arrows indicate the shift from the original to the constrained or expanded setting.}
        \label{fig:constrained_expanded_naive}
    \end{minipage}%
    \hfill
    \begin{minipage}[t]{0.36\textwidth}
        \centering
        \includegraphics[width=\textwidth]{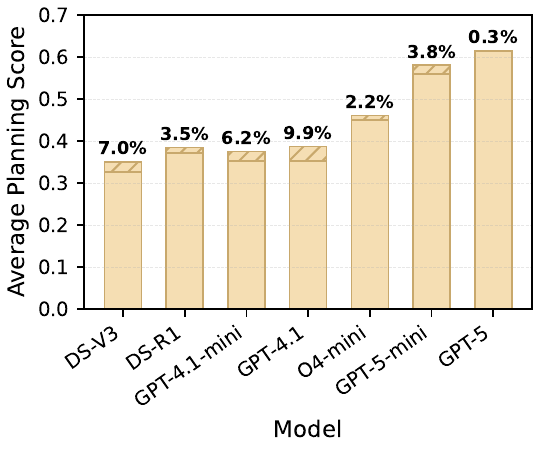}
        \caption{Impact of curated literature on average Planning Score across LLMs. Bars show baseline performance (solid) and the improvement with curated literature (hatched). DS denotes DeepSeek.}
        \label{fig:related_lit_improvement}
    \end{minipage}
\end{figure}

\paragraph{ReAct agent does not surpass simpler baselines.} Contrary to expectations, the ReAct agent does not improve performance compared to simpler baselines. We find that models sometimes struggle to intelligently filter retrieved information, thus incorporating irrelevant details that distract from the research idea. We show an example in Table~\ref{tab:react_knowledge_conflict}. One possible reason is the knowledge conflict between parametric and retrieved knowledge in language models~\citep{knowledge_conflict_survey, Taming_Knowledge_Conflicts}.
An extended evaluation with a newer model and an agentic harness is included in Appendix~\S\ref{appen:icml2026_codex}.

\paragraph{Beyond the AI domain.} \change{To validate the generalization of our pipeline beyond the AI domain, we create a second benchmark from Nature Mental Health papers and evaluate multiple LLMs (Table~\ref{tab:nmh_results}). We observe similar performance patterns: GPT-5 achieves the highest scores, and the literature section remains the most challenging across all models.}

\subsection{Length vs. Quality Analysis}

 Because our evaluation rubric measures coverage of key elements in a research plan, longer outputs may gain an advantage by mentioning more rubric-relevant content. To examine this potential bias, we analyze the relationship between output length and Average Planning Score. 
We prompt models with a soft length constraint (\eg ``\emph{Your response must be at most \texttt{<upper>} words.}''). We find that this simple approach effectively controls LLM's verbosity.

When output length is controlled (Figure~\ref{fig:constrained_expanded_naive}, left), GPT-5 and GPT-5-mini still achieve the highest Planning Scores. In the expanded-length setting (Figure~\ref{fig:constrained_expanded_naive}, right), we observe that both GPT-4.1 and o4-mini benefit from producing longer outputs. Despite these gains, GPT-5 consistently remains the top performer, suggesting that its advantage comes from higher plan quality.

\subsection{Enhanced Context Experiment} 
We hypothesize that ReAct agents underperform due to low-quality retrievals (see Appendix Table~\ref{tab:react_knowledge_conflict}). To test this, we provide models with up to three carefully curated, directly relevant papers, reflecting real-world scenarios where researchers begin with known references.
As shown in Figure~\ref{fig:related_lit_improvement}, curated literature leads to consistent gains across all models, with improvements ranging from 0.3\% to 9.9\%. Per-section results are reported in Table~\ref{tab:baseline_related} (Appendix \S\ref{appen:enhanced_context_experiment}).
Mid-tier models (\eg GPT-4.1) benefit most, indicating that external grounding compensates for weaker prior knowledge.

\subsection{Potential Training Strategy}
A natural progression towards improving the research planning capabilities of LLMs is to train models using idea–plan pairs derived from published research papers. We conduct supervised fine-tuning (SFT) of GPT-4.1-mini and GPT-4.1 on 2,000 idea–plan pairs extracted from randomly sampled ICML 2024 papers, and we evaluate their performance against the corresponding off-the-shelf models in a 1-shot scenario. As discussed in Appendix \S\ref{appen:sft_experiment}, the SFT models exhibit a decrease in overall performance (Table~\ref{tab:sft_comparison}). We find that the obtained fine-tuned models tend to hallucinate more, especially in the literature review sections. We offer some potential explanations and suggest future directions of research in Appendix \S\ref{appen:sft_experiment}.

\section{Conclusion}
We conducted an investigation into automating the conversion of scientific ideas into research plans, assessing how various LLMs handle this task. We proposed a framework that leverages published academic papers to extract paired ideas and plans for model assessment. Through comparative analysis of multiple LLMs using standard and agentic frameworks, we established a foundation for advancement in automated research planning.

\section{Ethics Statement}
Our framework operates on publicly available research papers and adheres to fair-use principles. Generated plans are intended for research analysis, not for automated publication or deployment. We recognize the potential misuse of AI systems for producing unverified or plagiarized scientific outputs and emphasize that our work aims to benchmark research planning capability, not to replace human scientific judgment. We also ensure that no personally identifiable or sensitive data is used in our datasets.

\bibliography{idea2plan}
\bibliographystyle{colm2026_conference}

\appendix
\section{Full Example of Jailbreak-Tax~\citep{jailbreak_tax} paper's Rubric and Research Plan}\label{appen:jailbreak_tax_example}

We present the full example in Table~\ref{tab:main_paper_example_rubric_etc}, including the grading rubric for the Jailbreak-Tax paper and the research plan generated by GPT-5~\citep{jailbreak_tax}. At the end of the paper, Tables~\ref{tab:rubric_jailbreak_tax_part1}--\ref{tab:rubric_jailbreak_tax_part4} show the complete evaluation rubric across four parts, while Tables~\ref{tab:research_plan_gpt5_part1}--\ref{tab:research_plan_gpt5_part3} present the full research plan generated by GPT-5.

\section{Dataset Construction}

\subsection{Research Plan and Idea Extraction}
\label{appen:dataset_plan_idea}
We provide the complete prompt templates used for extracting research plans and ideas in our dataset construction pipeline:
\begin{itemize}
\item \textit{Research Plan Template:} The Research Plan Template (Table~\ref{tab:research_plan_template}) defines the structured format for generating research plans, covering introduction, literature review, methods, experimental design, and resource considerations.
\item \textit{Idea Extraction:} The Research Idea Extraction Prompt (Table~\ref{tab:idea_extraction}) transforms paper abstracts into concise, first-person research ideas that focus on the proposed approach rather than results.
\item \textit{Research Plan Extraction:} The Research Plan Extraction Prompt (Table~\ref{tab:plan_extraction}) extracts detailed research plans from full papers following the Research Plan Template.
\end{itemize}

We also note that research ideas in our dataset vary in verbosity (see Table~\ref{tab:idea_verbosity_examples}). Our work focuses on research planning from a given idea, which differs from idea generation approaches~\citep{kumar2025can,hu2024nova} where the research idea itself is the output.

\subsection{Rubric Generation}
\label{appen:dataset_rubric}
The Rubric Generation Prompt (Table~\ref{tab:rubric_generation}) instructs the model to create evaluation criteria in JSON format for assessing research plans. Since this prompt is long, we separate it into two components that are plugged into the placeholders:
\begin{itemize}
\item \textit{Section-by-Section Guidance:} (Table~\ref{tab:section_guidance}) provides detailed instructions for generating consistent rubric questions across all sections.
\item \textit{Low-quality vs. High-quality Question Examples:} (Table~\ref{tab:poor_vs_better}) illustrates common pitfalls and demonstrates how to formulate clear, generalizable evaluation criteria.
\end{itemize}

\begin{table}[h]
\centering
\caption{Research Plan Template}
\label{tab:research_plan_template}
\tiny
\fbox{\begin{tabular}{p{0.93\textwidth}}
\textbf{Research Plan Template} \\
\\
\#\# 1. Introduction \\
\\
\#\#\# 1.1 Background \\
Describe the current limitations or challenges in the field and why they matter now. This should be one paragraph. \\
\\
\#\#\# 1.2 Primary Objectives \\
List specific, measurable goals that define project success. These should align with the contributions typically outlined in a paper's introduction. \\
\\
\#\#\# 1.3 Research Questions \\
State the core research questions for this research project. \\
\\
\#\# 2. Key Literatures \\
Identify key related works and relevant domains that inform your research. Begin by listing a few key domains, and for each cited work, briefly explain why it is important to your current research plan. \\
\\
\#\# 3. Methods \\
Describe the core techniques, model architectures, and/or training strategies to be used. This should be aligned with the method section in a paper. \\
\\
\#\# 4. Initial Experimental Design \\
Describe the high-level design of your first experiments. This should align with the experiments section of a paper. \\
\\
\#\# 5. Resources, Compliance, and Ethical Considerations \\
\\
\#\#\# 5.1 Resource Requirements \\
List and estimate the resources needed to carry out your research. Focus on the most informative metrics that impact the budget and feasibility, such as GPU hours, token usage for API calls, and human annotation costs. \\
\\
\#\#\# 5.2 Ethical and Compliance Considerations \\
Address data privacy, safety, and approval needs. \\
\end{tabular}}
\end{table}

\begin{table}[h]
\centering
\caption{Research Idea Extraction Prompt}
\label{tab:idea_extraction}
\tiny
\fbox{\begin{tabular}{p{0.93\textwidth}}
\textbf{Research Idea Extraction Prompt} \\
\\
Extract the core research idea from a paper abstract. Transform the abstract into a concise, intuitive research idea that focuses on the proposed approach, not the results. \\
\\
\#\# Task \\
Convert the abstract into a research idea using first-person language that captures: \\
- The problem we aim to solve \\
- Our proposed method/approach/dataset, etc. \\
- What makes it novel \\
\\
\#\# Instructions \\
- Remove the proposed method or dataset names (e.g., replace "HumanEval" with "a code generation benchmark") \\
- Remove experimental results and performance claims \\
- Keep the core technical approach and methodology \\
- Use first-person planning language (We propose, We aim, We will) \\
- Stay concise and intuitive \\
\\
\#\# Example \\
\\
**Input Abstract:** \\
"Scientific literature understanding is crucial for extracting targeted information and garnering insights, thereby significantly advancing scientific discovery. Despite the remarkable success of Large Language Models (LLMs), they face challenges in scientific literature understanding, primarily due to (1) a lack of scientific knowledge and (2) unfamiliarity with specialized scientific tasks. To develop an LLM specialized in scientific literature understanding, we propose a hybrid strategy that integrates continual pre-training (CPT) and supervised fine-tuning (SFT), to simultaneously infuse scientific domain knowledge and enhance instruction-following capabilities for domain-specific tasks. In this process, we identify two key challenges: (1) constructing high-quality CPT corpora, and (2) generating diverse SFT instructions. We address these challenges through a meticulous pipeline, including PDF text extraction, parsing content error correction, quality filtering, and synthetic instruction creation. Applying this strategy, we present a suite of LLMs: SciLitLLM, specialized in scientific literature understanding. These models demonstrate promising performance on scientific literature understanding benchmarks." \\
\\
**Output Research Idea:** \\
We propose a hybrid strategy that integrates continual pre-training (CPT) and supervised fine-tuning (SFT) to develop LLMs specialized in scientific literature understanding, addressing the challenges of lack of scientific knowledge and unfamiliarity with specialized scientific tasks. We aim to develop a pipeline for constructing high-quality CPT corpora and generating diverse SFT instructions through PDF text extraction, parsing error correction, and quality filtering. \\
\\
\#\# Input Abstract \\
\{\{ABSTRACT\}\} \\
\\
\#\# Output \\
Write the research idea using first-person style, focusing on the approach and novelty while removing experimental results and work names. \\
\end{tabular}}
\end{table}

\begin{table}[h]
\centering
\caption{Research Plan Extraction Prompt}
\label{tab:plan_extraction}
\tiny
\fbox{\begin{tabular}{p{0.93\textwidth}}
\textbf{Research Plan Extraction Prompt} \\
\\
Given the following research paper, extract a detailed research plan. Here is the definition of a research plan with some examples. Do not include the definition or any of the examples in your final output. \\
\\
// begin of research plan definition with examples \\
\\
\{\{RESEARCH\_PLAN\_DEFINITION\}\} \\
\\
// end of research plan definition \\
\\
**Additional notes when extracting the research plan:** \\
\\
- Do not include any information that is not explicitly stated in the paper. When listing items (metrics, datasets, models, techniques), include every such item mentioned in the paper. Never add details that is similar to the examples provided in the research plan definition, but not mentioned in the paper. \\
\\
- **Primary Objectives:** Identify the main goals of the research based on the Introduction section. Limit to a maximum of 5 objectives unless the paper clearly defines more. Focus on the most emphasized goals. \\
\\
- **Research Questions:** Ensure each question is distinct and non-overlapping. \\
\\
- **Related Literature:** Focus on identifying the most **foundational and influential** previous works that directly inspired this research or provide **key baselines, datasets, or theoretical foundations**. For each important work: \\
~~- Extract the full title and explain its specific contribution to the current research \\
~~- **Pay special attention to papers cited and discussed in the Methods and Experiments sections** - these are often the most critical works as they represent direct comparisons, baseline methods, or core datasets used \\
~~- Prioritize works that the authors: \\
~~~~- Compare their approach directly against (baseline methods) \\
~~~~- Build upon or extend (foundational methods) \\
~~~~- Use for evaluation (benchmark datasets) \\
~~~~- Cite as inspiration for their core methodology \\
~~- Explain how each cited work contributes - whether as comparative baselines, theoretical insights, datasets, or methodological foundations \\
~~- Focus on works discussed in detail rather than brief mentions \\
\\
- **Methodology:** Provide a detailed and logically structured description of the methods used. Ensure alignment with the stated objectives and research questions. \\
\\
- **Resource Requirements:** Extract any explicitly stated resource requirements mentioned in the paper (e.g., minimum number of GPUs needed, total GPU hours used, total number of tokens consumed, costs related to human annotation). Report only the order of magnitude rather than exact numbers. All resource-related details mentioned in the paper must be extracted. Do not make any guess on the resources if the paper does not specify resources. \\
\\
- **Ethics:** **Only include ethical considerations that are directly discussed in the paper.** Avoid speculative or generic concerns. Keep this section concise and focused. \\
\\
- In the research plan, use action-oriented, future-tense language that describes what will be done, not what was found. \\
\\
Here is the paper: \\
\\
// begin of the paper \\
\\
\{\{FULL\_PAPER\_TEXT\}\} \\
\\
// end of the paper \\
\\
Now give the extracted research plan, according to the research plan definition. \\
\end{tabular}}
\end{table}

\begin{table}[h]
\centering
\caption{Rubric Generation Prompt (Part 1: Instructions \& JSON Format)}
\label{tab:rubric_generation}
\tiny
\fbox{\begin{tabular}{p{0.93\textwidth}}
You are given a research idea and a plan for executing on that idea. Your task is to create a grading rubric in JSON format that can be used to evaluate other research plans on how well they develop the same research idea. The grading rubric will be used to assess and score alternative research plans that attempt to address the same research question. \\
\\
Here is the definition of a research plan: \\
\\
// begin of research plan definition with one example \\
\\
\{\{RESEARCH\_PLAN\_DEFINITION\}\} \\
\\
// end of research plan definition with one example \\
\\
\#\# Instructions \\
\\
1. Generalize the Criteria. \\
\\
Follow these steps in order: \\
\\
Step 1: For each section of the research plan, define what constitutes a high-quality research plan when developing the research idea. \\
\\
Step 2: Verify whether the input plan contains these quality characteristics. \\
\\
Step 3: Generate a list of yes/no questions that can be used to grade other research plans against this one. \\
- If a specific term, concept, or methodology appears in the research plan but NOT in the original research idea, do not require it in the rubric questions. \\
- Judge based on intent and conceptual alignment rather than specific terminology - if a paper proposes a specific term for something, other research plans should not be required to use that exact term, but should include things with similar intent. \\
- The rubric questions should capture the fundamental criteria that any research plan generated from this research idea should have. Do not ask about anything that is unique or special to this particular research plan. \\
\\
2. Section-by-Section Guidance. \\
\\
\{\{SECTION\_BY\_SECTION\_GUIDANCE\}\} \\
\\
3. Structure the Rubric in JSON Format. \\
\\
Your rubric should be organized as a JSON object with the major sections of a research plan as top-level keys. For each section, list the key elements to check for, starting from high-level concepts (e.g., overall goals) down to specific details (e.g., dataset types, model architectures, evaluation metrics). Each question should be a JSON object with the question text. The rubric must follow this exact JSON structure: \\
\\
\texttt{\{} \\
\texttt{~~"sections": \{} \\
\texttt{~~~~"Section Name": \{} \\
\texttt{~~~~~~"subsections": \{} \\
\texttt{~~~~~~~~"Subsection Name": \{} \\
\texttt{~~~~~~~~~~"questions": [} \\
\texttt{~~~~~~~~~~~~\{} \\
\texttt{~~~~~~~~~~~~~~"question": "Question text here?"} \\
\texttt{~~~~~~~~~~~~\}} \\
\texttt{~~~~~~~~~~]} \\
\texttt{~~~~~~~~\}} \\
\texttt{~~~~~~\}} \\
\texttt{~~~~\}} \\
\texttt{~~\}} \\
\texttt{\}} \\
If a section has no subsections and contains questions directly, use this format: \\
\\
\texttt{\{} \\
\texttt{~~"sections": \{} \\
\texttt{~~~~"Section Name": \{} \\
\texttt{~~~~~~"questions": [} \\
\texttt{~~~~~~~~\{} \\
\texttt{~~~~~~~~~~"question": "Question text here?"} \\
\texttt{~~~~~~~~\}} \\
\texttt{~~~~~~]} \\
\texttt{~~~~\}} \\
\texttt{~~\}} \\
\texttt{\}} \\
\end{tabular}}
\end{table}

\begin{table}[h]
\centering
\caption{Rubric Generation Prompt (Part 2: Examples \& Output)}
\label{tab:rubric_generation_part2}
\tiny
\fbox{\begin{tabular}{p{0.93\textwidth}}
\#\# Examples \\
\\
Here is an example of a rubric for a different plan: \\
\\
// begin of research plan rubric example \\
\\
\{\{RESEARCH\_PLAN\_RUBRIC\_EXAMPLE\}\} \\
\\
// end of research plan rubric example \\
\\
\{\{POOR\_VS\_BETTER\_EXAMPLES\}\} \\
\\
\#\# Output \\
\\
Now, here is the research plan and the research idea. Your task is to generate a structured grading rubric in JSON format: \\
\\
Here is the research idea: \\
\\
// begin of research idea \\
\\
\{\{RESEARCH\_IDEA\}\} \\
\\
// end of research idea \\
\\
// begin of research plan \\
\\
\{\{RESEARCH\_PLAN\}\} \\
\\
// end of research plan \\
\\
**IMPORTANT**: Your output must be valid JSON following the exact structure specified above. Do not include any text before or after the JSON object. Return only the JSON rubric. \\
\end{tabular}}
\end{table}

\begin{table}[h]
\centering
\caption{Section-by-Section Guidance for Rubric Generation}
\label{tab:section_guidance}
\tiny
\fbox{\begin{tabular}{p{0.93\textwidth}}
\textbf{Section-by-Section Guidance} \\
\\
Use the following guidance to ensure consistency and completeness across all rubric sections: \\
\\
- Introduction - Background: Write rubric questions that assess whether a plan clearly identifies the motivation and current limitations in the field. \\
\\
- Introduction - Primary Objectives: Write rubric questions that assess whether a plan defines specific, measurable goals. \\
\\
- Introduction - Research Questions: Write rubric questions that determine whether a plan articulates focused and relevant research questions. \\
\\
- Key Literature: Write rubric questions that check whether a plan cites key related work. This is the only section where in-depth discussion of related literature should occur. Group papers by domain. For each paper mentioned in the research plan, create one citation question using the format: "Does the plan cite [insert the paper title here] or similar work on [specific topic]?" \\
\\
Example: If the plan mentions "Attention Is All You Need", the question should be: "Does the plan cite the paper (Attention Is All You Need) or similar work on transformer architectures?" \\
\\
It's important not to require exact citation of the specific paper title. The paper title in the question is just an example. Focus on whether the plan cites any work that serves the same purpose or addresses the same topic, not whether it cites that exact paper. \\
\\
- Methods: Write rubric questions that assess whether a plan outlines a sound technical approach. Do not require an exact method match unless it is the only viable option. Generate questions that capture the high-level requirements for this plan. \\
\\
- Initial Experimental Design: Write rubric questions that assess whether a plan includes a clear and complete experimental setup. When you mention exact datasets or models, make sure to mention them as examples and put them in parentheses. \\
\\
- Resource Requirements: Write rubric questions that assess whether a research plan provides a realistic estimate of the resources required. \\
\\
- Ethical and Compliance Considerations: Write rubric questions that assess whether a plan addresses important ethical and legal responsibilities. \\
\end{tabular}}
\end{table}

\begin{table}[h]
\centering
\caption{Poor vs. Better Question Examples for Rubric Generation}
\label{tab:poor_vs_better}
\tiny
\fbox{\begin{tabular}{p{0.93\textwidth}}
\textbf{Poor vs. Better Question Examples} \\
\\
Example 1: \\
Poor: "Does the plan use task rewording or transformation (e.g., EvilMath) to trigger safety mechanisms while preserving semantic fidelity?" \\
Better: "Does the plan describe how evaluation tasks will be adapted to test safety mechanisms (e.g., using rewording)?" \\
Reason: "Semantic fidelity" is not defined and unclear. Requiring "use task rewording" is too specific, there could be other ways. \\
\\
Example 2: \\
Poor: "Does the plan propose analysis of the effect of model scale and attack type on the jailbreak tax?" \\
Better: "Does the plan propose analysis of the effect of model scale on performance?" and "Does the plan propose analysis of the effect of attack type on performance?" \\
Reason: You are asking two things, should be split into two questions. \\
\\
Example 3: \\
Poor: "Does the plan include a fine-tuning-based jailbreak using legitimate QA pairs?" \\
Better: "Does the plan include a fine-tuning-based jailbreak approach?" \\
Reason: "Legitimate QA pairs" is not clear in this question's context. \\
\\
Example 4: \\
Poor: "Does the plan define a clear metric for jailbreak success rate?" \\
Better: "Does the plan define a clear metric for evaluating attack effectiveness (e.g., success rate for jailbreak)?" \\
Reason: "Jailbreak success rate" is a specific term from the paper that the agent generating the plan may not capture, while "attack effectiveness" is more general. \\
\\
Example 5: \\
Poor: "Does the plan include visualizations of results across alignment techniques, models, and tasks?" \\
Better: "Does the plan propose analysis of results across alignment techniques, models, and tasks?" \\
Reason: Research plans describe what will be done, not the final outputs like visualizations. \\
\end{tabular}}
\end{table}

\begin{table}[h]
\centering
\small
\caption{Examples of research ideas with varying verbosity from Idea2Plan dataset.}
\label{tab:idea_verbosity_examples}
\begin{tabular}{p{0.15\textwidth}p{0.08\textwidth}p{0.70\textwidth}}
\toprule
\textbf{Paper} & \textbf{Word Count} & \textbf{Research Idea} \\
\midrule
Counterfactual Graphical Models: Constraints and Inference & 50 & We propose an ancestral graphical representation that unifies multiple hypothetical interventions into a single structure, enabling us to read counterfactual independences soundly and completely via d-separation. We then develop a counterfactual calculus—three transformation rules grounded in the graph's structural constraints—that extends the principles of interventional do-calculus to systematic counterfactual reasoning. \\
\midrule
Benign Samples Matter! Fine-tuning On Outlier Benign Samples Severely Breaks Safety & 63 & We propose a red-teaming framework that treats safety degradation during fine-tuning as an outlier detection problem: we will detect the small subset of samples in benign datasets that most undermine model alignment, then fine-tune LLMs exclusively on those outliers to expose hidden safety vulnerabilities. This approach reveals that seemingly harmless data can disproportionately compromise safety and underscores the need for stronger fine-tuning safeguards. \\
\midrule
Is Complex Query Answering Really Complex? & 90 & We propose to revisit complex query answering on knowledge graphs by first showing that most existing benchmark queries can be solved via single-edge link prediction, which masks true reasoning challenges. We aim to construct a new suite of queries that cannot be reduced to simple link predictions, instead requiring genuine multi-hop, intersection, and compositional reasoning over incomplete graphs. To do this, we will systematically filter out trivial queries and generate diverse, irreducible query structures that mirror real-world knowledge graph complexities, providing a more faithful evaluation framework for future CQA methods. \\
\midrule
Blink of an eye: a simple theory for feature localization in generative models & 91 & We propose to develop a unifying theoretical framework based on stochastic localization to explain why generative models suddenly shift behavior in narrow "critical windows" during inference. We aim to show that, as generation progresses, the model's distribution naturally concentrates onto a small sub-population, triggering abrupt changes in outputs. We will formulate this localization phenomenon with minimal distributional assumptions so that it applies both to sequence-based generation and to continuous diffusion-style processes, derive tighter quantitative bounds than prior analyses, and rely only on basic mathematical tools to make the theory broadly accessible. \\
\bottomrule
\end{tabular}
\end{table}

\section{Expert Evaluation of Reference Research Plan and Rubric}\label{appen:expert_papers}

Eight papers were selected by the expert evaluators, each from their respective research domain. The papers span diverse areas of AI and Machine Learning (ML) research:

\begin{itemize}
    \item \textbf{AI4Science.} The AI Scientist~\citep{ai-scientist}
    \item \textbf{Natural Language Processing and AI Agents.} GEPA~\citep{GEPA}, PaSa~\citep{PaSa}, Tree-of-Debate~\citep{Tree-of-Debate}
    \item \textbf{Graph ML.} Equivariance Everywhere All At Once~\citep{Equivariance}
    \item \textbf{Computer Vision.} UGround~\citep{GUI_agent}, AttWarp~\citep{Constructive_Distortion}.
    \item \textbf{Theoretical ML.} Mind the Gap~\citep{Mind_the_Gap}
\end{itemize}

\subsection{Guideline to Experts}

We provide comprehensive guidelines to expert annotators for evaluating both the LLM-generated research plans and rubric questions. The complete evaluation guidelines are included in Table~\ref{tab:evaluation_guidelines_part1} and Table~\ref{tab:evaluation_guidelines_part2}, which detail the rating scales and evaluation criteria for each section.

\begin{table}[h]
\centering
\caption{Human Evaluation Guidelines: Research Plan and Rubric Assessment, Part 1}
\label{tab:evaluation_guidelines_part1}
\tiny
\fbox{\begin{tabular}{p{0.93\textwidth}}
\textbf{What You're Evaluating} \\
\\
Hi! We're excited to have your help evaluating LLM-generated content using the 1-5 scale below. Your expert feedback is incredibly valuable to us and will help improve our research! \\
\\
Paper: <PAPER TITLE AND LINK PLACEHOLDER> \\
\\
You will evaluate TWO pieces of LLM-generated content: \\
1. LLM-Generated Research Plan from Full Paper - How well does the AI-extracted plan capture the original paper's content? \\
2. LLM-Generated Rubric Questions - How good are the AI-generated evaluation questions for assessing research plans? \\
\\
\hline
\\
\textbf{Definition of a Research Plan} \\
\\
<Placeholder for the definition of research plan, see Section X> \\
\\
\hline
\\
\textbf{PART 1: LLM-GENERATED RESEARCH PLAN FROM FULL PAPER TO EVALUATE} \\
\\
Instructions: The content below was automatically generated by an AI system from the research paper. In the next section we will provide guidelines for you to evaluate this. \\
\\
<INSERT GENERATED RESEARCH PLAN HERE> \\
\\
\hline
\\
\textbf{PART 2: YOUR EVALUATION OF THE RESEARCH PLAN} \\
\\
Instructions: Please rate how accurately the LLM captured the content from the original paper. \\
\\
\textbf{Rating Scale:} \\
$\bullet$ 1 = Major issues. The research plan fails to capture the key aspects of the evaluation criteria for the respective sections. \\
$\bullet$ 2 = Significant problems. The research plan only partially addresses the evaluation criteria and overlooks several important aspects. \\
$\bullet$ 3 = Average. The research plan covers the main criteria but contains notable gaps or inaccuracies compared to what the authors wrote in the paper. \\
$\bullet$ 4 = Acceptable with some issues. The research plan addresses most of the evaluation criteria well with only minor issues or omissions. \\
$\bullet$ 5 = Well done overall. The research plan is generally well-constructed and comprehensive, needing only minor adjustments. \\
\\
\textbf{Section 1: Introduction (Background, Objectives, Research Questions)} \\
Rate from 1-5: \_\_\_ \\
Evaluation Criteria: \\
$\bullet$ Does the background accurately reflect the paper's motivation and context? \\
$\bullet$ Are the objectives correctly extracted from the paper's stated goals? \\
$\bullet$ Do the research questions in the plan match those addressed in the original paper? \\
Comments: \_\_\_\_\_\_\_\_\_\_\_\_\_\_\_\_ \\
\\
\textbf{Section 2: Key Literature} \\
Rate from 1-5: \_\_\_ \\
Evaluation Criteria: \\
$\bullet$ Does it include the key references mentioned in the original paper? \\
$\bullet$ Are any important citations from the paper missing or misrepresented? \\
Comments: \_\_\_\_\_\_\_\_\_\_\_\_\_\_\_\_ \\
\\
\textbf{Section 3: Methods} \\
Rate from 1-5: \_\_\_ \\
Evaluation Criteria: \\
$\bullet$ Do the proposed methods accurately reflect the paper's methodology? \\
$\bullet$ Are the key technical details correctly extracted from the paper? \\
$\bullet$ Are any key methodological components from the paper missing? \\
Comments: \_\_\_\_\_\_\_\_\_\_\_\_\_\_\_\_ \\
\\
\textbf{Section 4: Experimental Design} \\
Rate from 1-5: \_\_\_ \\
Evaluation Criteria: \\
$\bullet$ Does the experimental design match the paper's evaluation setup? \\
$\bullet$ Are the datasets, metrics, and baselines correctly extracted? \\
$\bullet$ Are any important experimental details from the paper missing? \\
Comments: \_\_\_\_\_\_\_\_\_\_\_\_\_\_\_\_ \\
\\
\textbf{Section 5.1: Resources} \\
Rate from 1-5: \_\_\_ \\
Evaluation Criteria: \\
$\bullet$ Do the resource estimates align with what's described in the paper and appear realistic for the proposed research? \\
Comments: \_\_\_\_\_\_\_\_\_\_\_\_\_\_\_\_ \\
\\
\textbf{Section 5.2: Ethics} \\
Rate from 1-5: \_\_\_ \\
Evaluation Criteria: \\
$\bullet$ Are ethical considerations properly identified and addressed, including any mentioned in the paper? \\
Comments: \_\_\_\_\_\_\_\_\_\_\_\_\_\_\_\_
\end{tabular}}
\end{table}

\begin{table}[h]
\centering
\caption{Human Evaluation Guidelines: Research Plan and Rubric Assessment, Part 2}
\label{tab:evaluation_guidelines_part2}
\tiny
\fbox{\begin{tabular}{p{0.93\textwidth}}
\textbf{PART 3: YOUR EVALUATION OF THE RUBRIC QUESTIONS} \\
\\
Instructions: The rubric questions you are going to evaluate were automatically generated by an AI system. These questions are intended to evaluate research plans developed from the research idea below. (The given paper should be seen as the result of such research plan.) \\
\\
The research idea: <INSERT RESEARCH IDEA HERE> \\
\\
The rubric questions should: \\
$\bullet$ Assess whether the questions cover the essential parts of a research plan for the given idea that could have reasonably led to the given paper \\
$\bullet$ Focus on intent and conceptual alignment, rather than specific wording or terminology \\
\\
Please review the rubric questions below and rate their overall quality using the 1-5 scale. \\
\\
\textbf{Rating Scale:} \\
$\bullet$ 1 = Major issues. The rubric questions are irrelevant or do not address the key evaluation criteria for this section. \\
$\bullet$ 2 = Significant problems. The rubric questions have notable flaws or overlook several important aspects that should be assessed. \\
$\bullet$ 3 = Average. The rubric questions address some basic evaluation needs but could be improved. \\
$\bullet$ 4 = Acceptable with some issues. The rubric questions cover the essential evaluation needs but still have some gaps. \\
$\bullet$ 5 = Well done overall. The rubric questions are generally well-constructed and thorough, needing only minor adjustments. \\
\\
\hline
\\
\textbf{Section 1: Introduction Rubric Questions Quality} \\
<INSERT INTRODUCTION RUBRIC QUESTIONS HERE> \\
Rate the Introduction rubric questions from 1-5: \_\_\_ \\
Comments: \_\_\_\_\_\_\_\_\_\_\_\_\_\_\_\_ \\
\\
\textbf{Section 2: Key Literature Rubric Questions Quality} \\
<INSERT KEY LITERATURE RUBRIC QUESTIONS HERE> \\
Rate the Key Literature rubric questions from 1-5: \_\_\_ \\
Comments: \_\_\_\_\_\_\_\_\_\_\_\_\_\_\_\_ \\
\\
\textbf{Section 3: Methods Rubric Questions Quality} \\
<INSERT METHODS RUBRIC QUESTIONS HERE> \\
Rate the Methods rubric questions from 1-5: \_\_\_ \\
Comments: \_\_\_\_\_\_\_\_\_\_\_\_\_\_\_\_ \\
\\
\textbf{Section 4: Experimental Design Rubric Questions Quality} \\
<INSERT EXPERIMENTAL DESIGN RUBRIC QUESTIONS HERE> \\
Rate the Experimental Design rubric questions from 1-5: \_\_\_ \\
Comments: \_\_\_\_\_\_\_\_\_\_\_\_\_\_\_\_ \\
\\
\textbf{Section 5: Resources and Ethics Rubric Questions Quality} \\
<INSERT RESOURCES AND ETHICS RUBRIC QUESTIONS HERE> \\
Rate Resources rubric questions from 1-5: \_\_\_ \\
Rate Ethics rubric questions from 1-5: \_\_\_ \\
Comments: \_\_\_\_\_\_\_\_\_\_\_\_\_\_\_\_
\end{tabular}}
\end{table}

\subsection{Characteristics of Annotators}

We recruit eight volunteer annotators who are experts in AI. All annotators have specialization in AI and natural language processing. Our annotator team consists of researchers with the following demographic composition: 75\% from Asia, 12.5\% from the Middle East, and 12.5\% from the United States.  Each annotator independently evaluated assigned papers following the guideline. Our annotators are informed that their provided data would be used solely for research purposes in this study.

\section{LLM-based Judge}\label{appen:llm_judge}

We use an LLM-as-a-judge approach~\citep{llm_judge} to evaluate research plans against the generated rubrics, with the Rubric Evaluation Prompt (Table~\ref{tab:rubric_evaluation}) guiding the model to perform assessment with strict interpretation of rubric criteria. We use o4-mini (reasoning=high) throughout the study.
Because grading an entire plan in one API call can exceed input limits and cause failures, we instead evaluate each section separately—making five API requests per plan (one for each section)—and then aggregate the section scores to obtain the final overall score.

\begin{table}[h]
\centering
\caption{Rubric Evaluation Prompt (Part 1: Instructions \& Inputs)}
\label{tab:rubric_evaluation}
\tiny
\fbox{\begin{tabular}{p{0.93\textwidth}}
You are given a grading rubric and a new research plan. Your task is to evaluate the new plan against the rubric using a **strict interpretation**: only answer **Yes** if the plan explicitly satisfies the rubric question as written. Note that examples in parentheses (e.g., specific datasets, papers, methods) are for reference only. \\

\#\# Instructions \\
\\
1. You will be evaluating **ONE SPECIFIC SECTION** of the rubric at a time against the entire research plan. \\
\\
2. For the given section: \\
~~~- Traverse all levels of the section hierarchy. Rubric items may be nested (e.g., subsections → questions). \\
~~~- Evaluate each **leaf-level question** (i.e., the final bullet points that are actual rubric questions). \\
~~~- Answer **Yes** only if the research plan clearly and explicitly addresses the rubric question. \\
~~~- Answer **No** if the rubric question is not addressed, or if the answer is vague or only implied. \\
\\
~~~> For example: \\
~~~> - General references are not sufficient unless they fully preserve the original intent of rubric item. \\
\\
3. When rubric questions include examples in parentheses (for example, "e.g., dataset A, dataset B"), these are provided as reference examples to illustrate the type of content being asked about. Do NOT require the research plan to mention these specific examples to answer "Yes". \\
\\
~~~> For example: \\
~~~> - If the rubric asks "Does the plan use mathematical datasets (e.g., GSM8K, MATH)?", answer "Yes" if the plan uses any appropriate mathematical datasets, not just GSM8K or MATH specifically \\
\\
4. For each rubric question in the specified section: \\
~~~- **Question**: [rubric question] \\
~~~- **Answer**: Yes / No \\
~~~- **Explanation**: [brief justification] \\
\\
--- \\
\\
\#\# Inputs \\
\\
// begin of research plan rubric section to evaluate \\
\{\{RESEARCH\_PLAN\_RUBRIC\_SECTION\}\} \\
// end of research plan rubric section \\
\\
// begin of full research plan \\
\{\{RESEARCH\_PLAN\}\} \\
// end of full research plan \\
\\
// section being evaluated \\
\{\{SECTION\_NAME\}\} \\
\end{tabular}}
\end{table}

\begin{table}[h]
\centering
\caption{Rubric Evaluation Prompt (Part 2: Output Format \& Notes)}
\label{tab:rubric_evaluation_part2}
\tiny
\fbox{\begin{tabular}{p{0.93\textwidth}}
\#\# Output Format \\
\\
Please return the evaluation for the specified section in the following structured JSON format: \\
\\
\texttt{\{} \\
\texttt{~~"section\_name": "\{\{SECTION\_NAME\}\}",} \\
\texttt{~~"evaluation": \{} \\
\texttt{~~~~"subsections": \{} \\
\texttt{~~~~~~"Subsection Title A": \{} \\
\texttt{~~~~~~~~"questions": [} \\
\texttt{~~~~~~~~~~\{} \\
\texttt{~~~~~~~~~~~~"question": "Rubric question 1",} \\
\texttt{~~~~~~~~~~~~"answer": "Yes" or "No",} \\
\texttt{~~~~~~~~~~~~"explanation": "Your explanation here"} \\
\texttt{~~~~~~~~~~\},} \\
\texttt{~~~~~~~~~~...} \\
\texttt{~~~~~~~~]} \\
\texttt{~~~~~~\},} \\
\texttt{~~~~~~...} \\
\texttt{~~~~\}} \\
\texttt{~~\}} \\
\texttt{\}} \\
\\
**Note**: If the section has no subsections and contains questions directly, use this format instead: \\
\\
\texttt{\{} \\
\texttt{~~"section\_name": "\{\{SECTION\_NAME\}\}",} \\
\texttt{~~"evaluation": \{} \\
\texttt{~~~~"questions": [} \\
\texttt{~~~~~~\{} \\
\texttt{~~~~~~~~"question": "Rubric question 1",} \\
\texttt{~~~~~~~~"answer": "Yes" or "No",} \\
\texttt{~~~~~~~~"explanation": "Your explanation here"} \\
\texttt{~~~~~~\},} \\
\texttt{~~~~~~...} \\
\texttt{~~~~]} \\
\texttt{~~\}} \\
\texttt{\}} \\
\\
--- \\
\\
\#\# Important JSON Formatting Notes: \\
- When mentioning dollar amounts, use "\$" not "\textbackslash\textbackslash\$" \\
- Avoid unnecessary escape characters in explanations \\
- Valid JSON escapes are: \textbackslash\textbackslash n \textbackslash\textbackslash t \textbackslash\textbackslash r \textbackslash\textbackslash b \textbackslash\textbackslash f \textbackslash\textbackslash" \textbackslash\textbackslash\textbackslash\textbackslash{} \textbackslash\textbackslash/ \\
- Do not escape regular punctuation like \$ or other symbols \\
\\
\#\# Important Notes \\
\\
- You are evaluating **only the section specified** in the SECTION\_NAME field \\
- Search through the **entire research plan** to find evidence for each rubric question \\
- Be thorough but focus only on the questions within the specified section \\
- This evaluation will be combined with evaluations of other sections to form a complete assessment \\
\end{tabular}}
\end{table}

\section{Baseline and Agent Design}\label{appen:all_prompt_baselines_react}

\subsection{Prompting Baselines}
\label{appen:prompting_baselines}

We design three prompting baselines for research plan generation:
\begin{itemize}
\item \textit{Naïve Baseline:} The Naïve Baseline Prompt (Table~\ref{tab:baseline_naive}) provides minimal instructions with only the research idea and template structure.
\item \textit{0-shot Baseline:} The 0-shot Baseline Prompt (Table~\ref{tab:baseline_0shot}) adds explicit generation instructions to guide the model through the planning process.
\item \textit{1-shot Baseline:} The 1-shot Baseline Prompt (Table~\ref{tab:baseline_1shot}) includes the same instructions as the 0-shot baseline, with one example research plan.
\end{itemize}

\subsection{ReAct Agent}
\label{appen:react_agent_prompt}

The ReAct Agent Prompt (Table~\ref{tab:react_agent}) implements an iterative reasoning framework that enables the agent to systematically gather information through tool use before generating the research plan~\citep{ReAct}. While we focus on ReAct as a widely-used single-agent framework, multi-agent approaches such as debate-based methods~\citep{liang2024encouraging} could be explored in future work. The agent has access to two tools specified in Table~\ref{tab:tool_specs}: \texttt{search\_papers} for finding relevant academic papers using Bing Custom Search, and \texttt{read\_paper} for retrieving and analyzing specific papers by their arXiv ID. These tool specifications (referred to as \texttt{\{\{TOOL\_SPECIFICATIONS\}\}} in the prompt) and tool names (referred to as \texttt{\{\{TOOL\_NAMES\}\}} in the prompt) are provided to the agent to enable systematic information gathering.

The \texttt{search\_papers} tool returns the top 10 most relevant papers, each including its arXiv ID, title, and abstract.
When the agent calls \texttt{read\_paper}, it uses the Paper Summarization Prompt (Table~\ref{tab:paper_summary_prompt}) to produce structured summaries of each retrieved paper. This prompt guides the model to extract a paper’s main contributions, key related literature, methods and techniques, and experimental design and results.

\begin{table}[h]
\centering
\caption{Naïve Baseline Prompt}
\label{tab:baseline_naive}
\tiny
\fbox{\begin{tabular}{p{0.93\textwidth}}
\textbf{Naïve Baseline Prompt} \\
\\
RESEARCH IDEA TO ANALYZE: \\
\\
\{\{RESEARCH\_IDEA\}\} \\
\\
Your task is to generate a detailed research plan based on the provided research idea. Below is the template you must follow to create the research plan: \\
\\
\#\# 1. Introduction \\
\\
\#\#\# 1.1 Background \\
\\
\#\#\# 1.2 Primary Objectives \\
\\
\#\#\# 1.3 Research Questions \\
\\
\#\# 2. Key Literatures \\
\\
\#\# 3. Methods \\
\\
\#\# 4. Initial Experimental Design \\
\\
\#\# 5. Resources, Compliance, and Ethical Considerations \\
\\
\#\#\# 5.1 Resource Requirements \\
\\
\#\#\# 5.2 Ethical and Compliance Considerations \\
\end{tabular}}
\end{table}

\begin{table}[h]
\centering
\caption{Zero-Shot Baseline Prompt}
\label{tab:baseline_0shot}
\tiny
\fbox{\begin{tabular}{p{0.93\textwidth}}
\textbf{Zero-Shot Baseline Prompt} \\
\\
RESEARCH IDEA TO ANALYZE: \\
\\
\{\{RESEARCH\_IDEA\}\} \\
\\
Your task is to generate a detailed research plan based on the provided research idea. Below is the template you must follow to create the research plan: \\
\\
\{\{RESEARCH\_PLAN\_TEMPLATE\}\} \\
\\
INSTRUCTIONS FOR RESEARCH PLAN GENERATION: \\
1. Analyze the research idea thoroughly from the provided research idea. \\
2. Generate a complete research plan following the EXACT template structure above. \\
3. Fill in all sections with relevant, specific content based on the research idea. \\
4. Draw upon your existing knowledge of the research area to provide context and background. \\
5. Include resource requirements for conducting the research. \\
\\
IMPORTANT NOTES: \\
- Base your plan on the provided research idea and your existing knowledge. \\
\end{tabular}}
\end{table}

\begin{table}[h]
\centering
\caption{One-Shot Baseline Prompt}
\label{tab:baseline_1shot}
\tiny
\fbox{\begin{tabular}{p{0.93\textwidth}}
\textbf{One-Shot Baseline Prompt} \\
\\
RESEARCH IDEA TO ANALYZE: \\
\\
\{\{RESEARCH\_IDEA\}\} \\
\\
Your task is to generate a detailed research plan based on the provided research idea. Below is the template you must follow to create the research plan: \\
\\
\{\{RESEARCH\_PLAN\_TEMPLATE\_WITH\_EXAMPLE\}\} \\
\\
INSTRUCTIONS FOR RESEARCH PLAN GENERATION: \\
1. Analyze the research idea thoroughly from the provided research idea. \\
2. Generate a complete research plan following the EXACT template structure above. \\
3. Fill in all sections with relevant, specific content based on the research idea. \\
4. Draw upon your existing knowledge of the research area to provide context and background. \\
5. Include resource requirements for conducting the research. \\
\\
IMPORTANT NOTES: \\
- **The examples provided in the research plan template are for reference only** - replace them with content specific to the given research idea. \\
- Base your plan on the provided research idea and your existing knowledge. \\
\end{tabular}}
\end{table}

\begin{table}[h]
\centering
\caption{ReAct Agent Prompt}
\label{tab:react_agent}
\tiny
\fbox{
\begin{minipage}{0.93\textwidth}
\raggedright
ReAct Agent Prompt

\vspace{4pt}
You are a research planning agent that generates comprehensive research plans using iterative reasoning.

\vspace{4pt}
RESEARCH IDEA: \{\{RESEARCH\_IDEA\}\}

\vspace{4pt}
Below is the definition of a research plan structure as well as one example.

\vspace{4pt}
IMPORTANT NOTE: The example provided is from a specific research plan about scientific literature understanding with LLMs. You should NOT be restricted to the specific settings, methods, datasets, or approaches mentioned in this example. Adapt all content to fit the specific research idea you are working on. Use your research knowledge and the information you gather through searches to create a plan that is most appropriate for the given research idea.

\vspace{4pt}
RESEARCH PLAN TEMPLATE AND GUIDELINES: \{\{RESEARCH\_PLAN\_DEFINITION\}\}

\vspace{4pt}
AVAILABLE TOOLS: \{\{TOOL\_NAMES\}\}

\vspace{4pt}
TOOL SPECIFICATIONS: \{\{TOOL\_SPECIFICATIONS\}\}

\vspace{4pt}
RESOURCE LIMITS: 
\begin{itemize}[leftmargin=*]
\item You have \{\{MAX\_TOTAL\_TOOLS\}\} total tool calls available across maximum \{\{MAX\_ITERATIONS\}\} iterations
\item \{\{CURRENT\_USAGE\_STATS\}\}
\end{itemize}

\vspace{4pt}
INSTRUCTIONS: Use the following ReAct format to systematically gather information and create a research plan.

\vspace{4pt}
IMPORTANT: Always begin with “Thought:” and explain your reasoning before taking any action.

\vspace{4pt}
FORMAT FOR EACH ITERATION:  
\texttt{Thought: [Your reasoning about what information you need]}  
\texttt{Action: [MUST be exactly one of: \{\{TOOL\_NAMES\}\}]}  
\texttt{Action Input: [specific query or input for the chosen tool]}

\vspace{4pt}
CRITICAL RULES:
\begin{enumerate}[leftmargin=*]
\item You MUST STOP after “Action Input:” — do NOT generate “Observation:” or any results
\item The system will execute your action and provide the real observation
\item Wait for the actual tool results before continuing
\item Use the EXACT tool names listed above. Available tools are: \{\{TOOL\_NAMES\}\}
\end{enumerate}

\vspace{4pt}
FINAL STEP (After gathering sufficient information):  
When you have collected enough information through your searches and reads, generate your final answer in this EXACT format:

\vspace{4pt}
\texttt{Thought: I now have sufficient information to create the research plan.}  
\texttt{Final Answer:}  
[Start your complete research plan here, following the template provided above. The plan should be comprehensive and detailed based on all the information you gathered through your tool calls.]

\vspace{4pt}
CRITICAL: You MUST include “Final Answer:” exactly as shown above. This is required for the system to recognize your final output.

\vspace{4pt}
EXAMPLE OF CORRECT FORMAT:  
\texttt{Thought: I need to find papers about attention mechanisms to understand current transformer architectures.}  
\texttt{Action: search\_papers}  
\texttt{Action Input: attention mechanisms transformer architectures}  
\texttt{\mbox{[STOP HERE - Wait for system to provide observation]}}

\vspace{4pt}
Key guidelines:
\begin{enumerate}[leftmargin=*]
\item Use targeted queries to find relevant information
\item Look for recent advances, existing methods, and gaps in the literature
\item Consider both technical approaches and evaluation methods
\item Each tool call should build upon previous findings
\item Choose the most appropriate tool for each information need
\item End with “RESEARCH PLAN COMPLETE” after your final answer
\item NEVER generate “Observation:” — always wait for the system response
\end{enumerate}

\vspace{4pt}
READING PAPERS: When you use the \texttt{read\_paper} action, you will receive a comprehensive summary of the paper content instead of the full text.

\vspace{4pt}
Begin your research planning process now. Start with a “Thought:” about what information you need to gather first.

\vspace{4pt}
REMEMBER: You must use the ReAct format. Do NOT attempt to create the research plan immediately. Start by thinking about what information you need, then use tools to gather that information systematically. STOP after each “Action Input:” and wait for the system’s observation.
\end{minipage}
}
\end{table}

\begin{table}[h]
\centering
\caption{Tool Specifications for ReAct Agent}
\label{tab:tool_specs}
\tiny
\fbox{\begin{tabular}{p{0.93\textwidth}}
1. search\_papers: Search for academic papers using Bing Custom Search \\
~~~- Input: A search query string for academic papers \\
~~~- Output: JSON with paper titles, abstracts, arXiv IDs, and publication details \\
~~~- Usage example: Action Input: machine learning attention mechanisms \\
\\
2. read\_paper: Read and analyze a specific academic paper by its arXiv ID \\
~~~- Input: An arXiv ID (e.g., "2301.12345") \\
~~~- Output: JSON with arXiv\_id and paper summary \\
~~~- Usage example: Action Input: 2301.12345 \\
\end{tabular}}
\end{table}

\begin{table}[h]
\centering
\caption{Paper Summarization Prompt}
\label{tab:paper_summary_prompt}
\tiny
\fbox{\begin{tabular}{p{0.93\textwidth}}
\texttt{summary\_prompt = f""" }\\
PAPER TO SUMMARIZE: \\
Title: \{paper\_title\} \\
ArXiv ID: \{arxiv\_id\} \\
\\
Paper Content: \\
\{paper\_content\} \\
\\
--- \\
\\
Create a comprehensive summary of this paper. Your summary should include: \\
\\
1. Main Contributions: Key findings and contributions of this paper \\
2. Key Related Literature: Important prior works referenced and how this paper builds upon or differs from them \\
3. Methods and Techniques: Approaches, algorithms, or methodologies used \\
4. Experimental Design and Results: How experiments were conducted, datasets used, evaluation metrics, and key results obtained \\
\\
SUMMARY: \\
\texttt{"""} \\
\end{tabular}}
\end{table}

\section{Full Experimental Results}
\label{appen:full_results}

Table~\ref{tab:mean_scores_full} and Table~\ref{tab:max_scores_full} present detailed section-wise results across all models and prompting settings. For each research idea, we run every configuration three times. The \textbf{mean} results report the average score over the three runs, while the \textbf{max} results report the best-performing run among them. Consistent trends appear across both metrics: GPT-5 achieves the highest scores under all prompting setups, followed by GPT-5-mini and o4-mini. 1-shot prompting provides a small but steady improvement over 0-shot, indicating that a single example helps models organize their research plans more coherently. In contrast, the ReAct setup does not outperform simpler prompting strategies.

\begin{table}[t]
\centering
\scriptsize
\setlength{\tabcolsep}{3pt}
\caption{Section-wise Planning Scores (\%) with \textbf{Mean} aggregation. Each value represents the mean accuracy across all papers in Idea2Plan Bench. Bold numbers indicate the highest score within each section across all baselines.}
\label{tab:mean_scores_full}
\begin{tabular}{l:rrrrr:r:rrrrr:r}
\toprule
\multirow{2}{*}{Model} & \multicolumn{6}{c:}{Naïve} & \multicolumn{6}{c}{0-shot} \\
\cmidrule(lr){2-7} \cmidrule(lr){8-13}
 & Intro & Lit & Met & Exp & Res/Eth & Avg & Intro & Lit & Met & Exp & Res/Eth & Avg \\
\midrule
Phi-4        & 29.0 & 5.6 & 20.2 & 10.9 & 30.9 & 19.3 & 28.8 & 8.2 & 20.9 & 10.7 & 43.1 & 22.3 \\
DeepSeek-V3  & 39.8 & 25.2 & 32.4 & 24.7 & 44.5 & 33.3 & 40.6 & 23.0 & 33.4 & 24.6 & 52.0 & 34.7 \\
DeepSeek-R1  & 40.6 & 27.1 & 37.0 & 28.7 & 50.3 & 36.7 & 41.3 & 24.5 & 37.2 & 29.7 & 56.8 & 37.9 \\
GPT-4.1-mini & 42.3 & 21.8 & 34.7 & 27.4 & 47.2 & 34.7 & 46.8 & 24.3 & 39.9 & 30.9 & 62.7 & 40.9 \\
GPT-4.1      & 40.7 & 22.4 & 34.2 & 28.9 & 48.7 & 34.9 & 44.0 & 23.7 & 34.9 & 29.2 & 60.4 & 38.4 \\
O4-mini      & 50.8 & 29.1 & 50.2 & 39.1 & 59.7 & 45.7 & 52.8 & 27.5 & 51.4 & 39.9 & 68.1 & 47.9 \\
GPT-5-mini   & 58.5 & 38.1 & 65.9 & 51.6 & 69.5 & 56.7 & 61.3 & 36.6 & 64.4 & 52.3 & 78.9 & 58.7 \\
GPT-5        & 60.9 & 47.7 & \textbf{70.9} & 56.7 & 73.8 & 62.0 & 63.2 & 43.9 & 69.6 & 56.9 & \textbf{81.2} & 63.0 \\
\midrule
\multirow{2}{*}{Model} & \multicolumn{6}{c:}{1-shot} & \multicolumn{6}{c}{ReAct} \\
\cmidrule(lr){2-7} \cmidrule(lr){8-13}
 & Intro & Lit & Met & Exp & Res/Eth & Avg & Intro & Lit & Met & Exp & Res/Eth & Avg \\
\midrule
Phi-4        & 28.6 & 9.2 & 20.1 & 12.9 & 45.1 & 23.2 & 24.2 & 11.8 & 15.5 & 9.9  & 31.7 & 18.6 \\
DeepSeek-V3  & 41.8 & 23.1 & 33.6 & 28.5 & 55.9 & 36.6 & 39.9 & 23.8 & 33.4 & 28.5 & 50.9 & 35.3 \\
DeepSeek-R1  & 40.3 & 25.3 & 35.7 & 30.4 & 57.1 & 37.8 & 38.7 & 23.1 & 33.2 & 27.9 & 52.0 & 35.0 \\
GPT-4.1-mini & 48.2 & 26.6 & 40.4 & 33.8 & 65.3 & 42.8 & 50.6 & 30.6 & 44.5 & 35.7 & 60.3 & 44.3 \\
GPT-4.1      & 45.2 & 24.0 & 38.1 & 31.7 & 65.2 & 40.8 & 50.6 & 35.6 & 44.9 & 40.5 & 65.6 & 47.4 \\
O4-mini      & 53.9 & 29.9 & 53.9 & 43.5 & 69.9 & 50.2 & 54.2 & 36.1 & 54.4 & 42.6 & 61.2 & 49.6 \\
GPT-5-mini   & 61.4 & 37.5 & 65.8 & 52.3 & 80.4 & 59.5 & 62.3 & 44.5 & 67.3 & 55.2 & 76.4 & 61.1 \\
GPT-5        & \textbf{63.2} & 44.2 & 70.2 & \textbf{57.4} & 81.0 & \textbf{63.2} & 61.4 & \textbf{48.1} & 68.4 & 55.7 & 75.9 & 61.9 \\
\bottomrule
\end{tabular}
\end{table}

\begin{table}[t]
\centering
\scriptsize
\setlength{\tabcolsep}{3pt}
\caption{Section-wise Planning Scores (\%) with \textbf{Max} aggregation. Each value reports the maximum accuracy across all generated plans per paper. Bold numbers indicate the highest score within each section across all baselines.}
\label{tab:max_scores_full}
\begin{tabular}{l:rrrrr:r:rrrrr:r}
\toprule
\multirow{2}{*}{Model} & \multicolumn{6}{c:}{Naïve} & \multicolumn{6}{c}{0-shot} \\
\cmidrule(lr){2-7} \cmidrule(lr){8-13}
 & Intro & Lit & Met & Exp & Res/Eth & Avg & Intro & Lit & Met & Exp & Res/Eth & Avg \\
\midrule
Phi-4        & 36.4 & 10.8 & 25.6 & 17.2 & 39.1 & 23.2 & 36.1 & 13.2 & 26.0 & 16.9 & 53.1 & 25.9 \\
DeepSeek-V3  & 46.9 & 32.2 & 39.9 & 31.8 & 53.6 & 37.4 & 47.7 & 30.8 & 41.5 & 32.1 & 61.0 & 39.2 \\
DeepSeek-R1  & 48.1 & 35.4 & 45.8 & 36.6 & 59.4 & 41.2 & 49.1 & 32.2 & 45.9 & 38.3 & 66.7 & 42.5 \\
GPT-4.1-mini & 49.9 & 30.2 & 41.9 & 35.1 & 57.2 & 39.0 & 54.3 & 32.6 & 47.7 & 39.4 & 72.9 & 45.2 \\
GPT-4.1      & 49.1 & 30.8 & 42.3 & 37.0 & 58.0 & 39.4 & 51.9 & 31.9 & 43.4 & 37.6 & 70.9 & 43.2 \\
O4-mini      & 59.0 & 38.5 & 59.3 & 46.9 & 70.1 & 50.5 & 61.0 & 35.6 & 60.1 & 48.3 & 77.7 & 52.2 \\
GPT-5-mini   & 66.3 & 48.2 & 73.4 & 60.4 & 77.7 & 61.4 & 69.8 & 46.0 & 72.9 & 61.8 & 86.9 & 63.4 \\
GPT-5        & 69.9 & 57.5 & \textbf{79.1} & 66.0 & 81.5 & 66.6 & 71.5 & 52.7 & 77.5 & \textbf{67.5} & \textbf{89.5} & 67.9 \\
\midrule
\multirow{2}{*}{Model} & \multicolumn{6}{c:}{1-shot} & \multicolumn{6}{c}{ReAct} \\
\cmidrule(lr){2-7} \cmidrule(lr){8-13}
 & Intro & Lit & Met & Exp & Res/Eth & Avg & Intro & Lit & Met & Exp & Res/Eth & Avg \\
\midrule
Phi-4        & 35.2 & 15.0 & 25.6 & 19.0 & 54.2 & 27.0 & 34.4 & 19.8 & 22.7 & 17.5 & 45.5 & 25.0 \\
DeepSeek-V3  & 49.0 & 30.6 & 41.7 & 36.2 & 64.8 & 41.2 & 48.2 & 32.4 & 42.1 & 36.6 & 60.6 & 40.3 \\
DeepSeek-R1  & 47.7 & 33.4 & 44.2 & 38.6 & 67.1 & 42.2 & 46.3 & 33.1 & 43.0 & 35.6 & 62.2 & 39.8 \\
GPT-4.1-mini & 57.8 & 36.1 & 52.9 & 46.3 & 76.9 & 49.1 & 58.6 & 41.4 & 54.1 & 45.3 & 70.5 & 50.0 \\
GPT-4.1      & 54.4 & 33.5 & 49.7 & 42.7 & 77.4 & 46.7 & 59.3 & 47.0 & 54.0 & 52.0 & 77.4 & 53.6 \\
O4-mini      & 63.0 & 38.9 & 65.5 & 55.8 & 81.6 & 55.8 & 63.5 & 47.2 & 65.4 & 52.5 & 72.2 & 55.9 \\
GPT-5-mini   & 69.8 & 47.0 & 74.7 & 63.2 & 88.6 & 64.4 & 70.9 & 55.4 & 77.1 & 66.6 & 84.2 & 66.2 \\
GPT-5        & \textbf{71.7} & 54.2 & 79.0 & 67.0 & 88.5 & \textbf{68.1} & 69.7 & \textbf{57.8} & 77.1 & 66.3 & 84.8 & 67.3 \\
\bottomrule
\end{tabular}
\end{table}

\section{Win Rate between LLMs}

\begin{figure}[t]
    \centering
    \begin{tabular}{cc}
        \includegraphics[width=0.47\textwidth, trim={20 25 20 30}, clip]{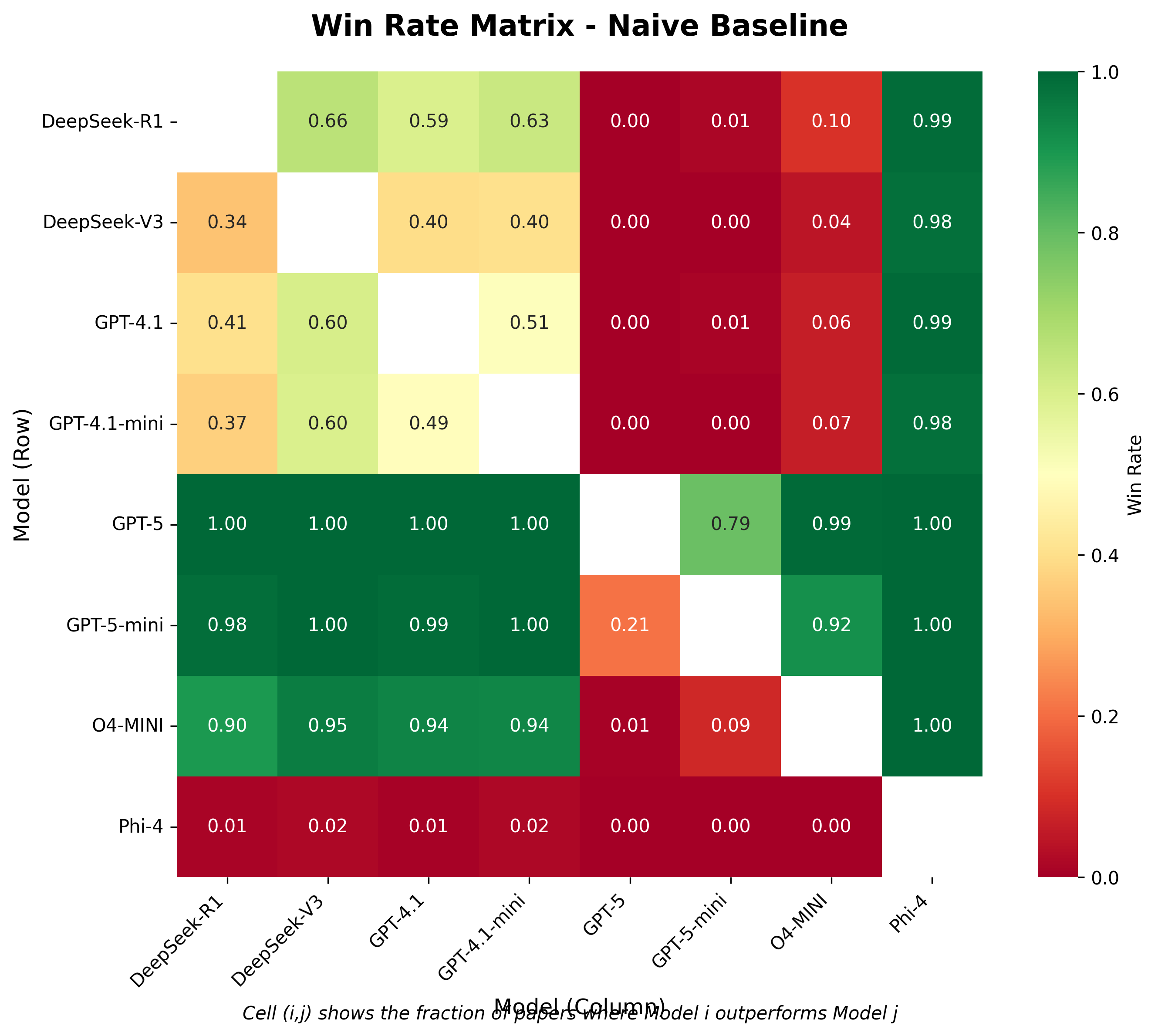} &
        \includegraphics[width=0.47\textwidth, trim={20 25 20 30}, clip]{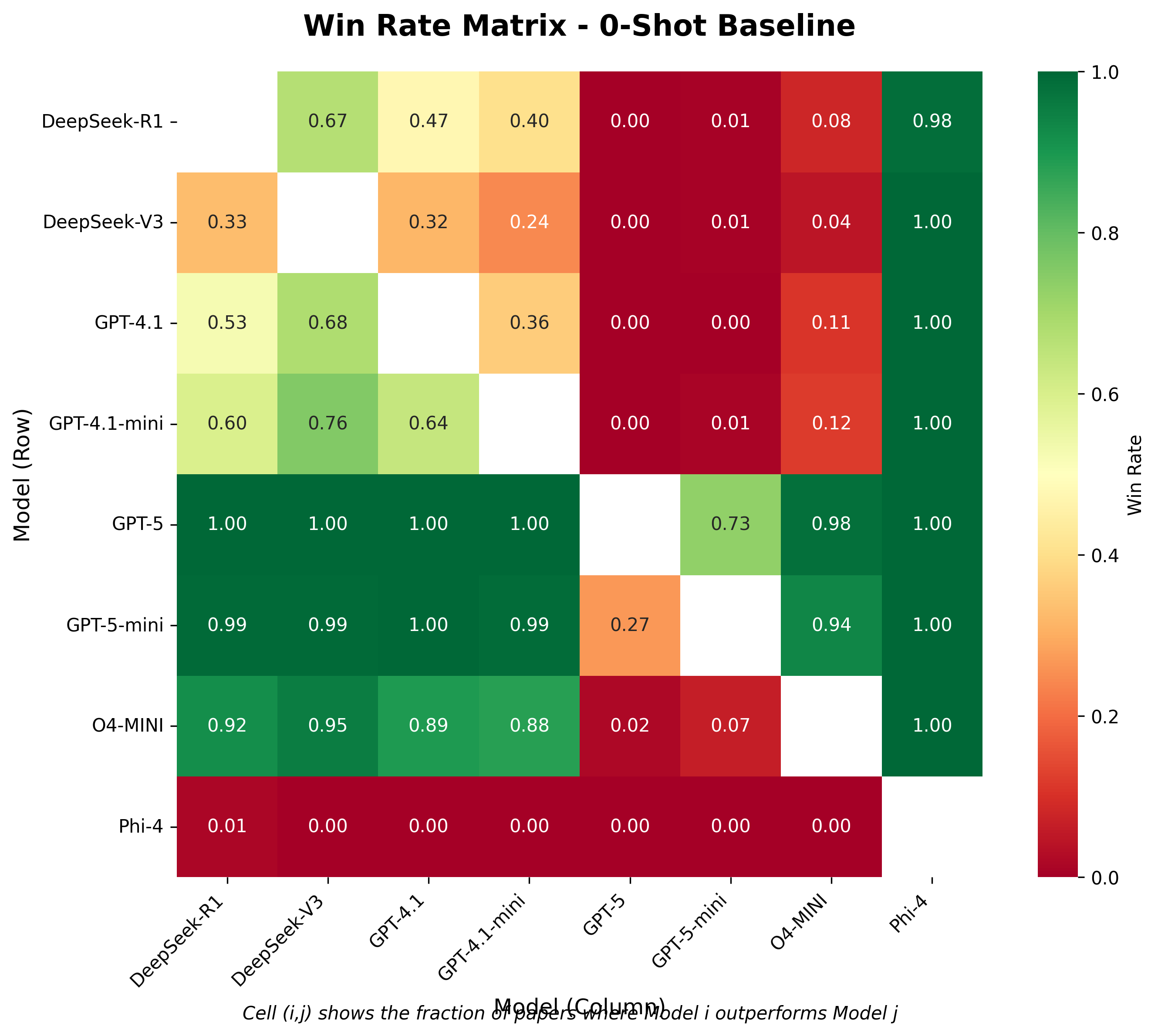} \\
        (a) Naïve & (b) 0-shot \\[6pt]
        \includegraphics[width=0.47\textwidth, trim={20 25 20 30}, clip]{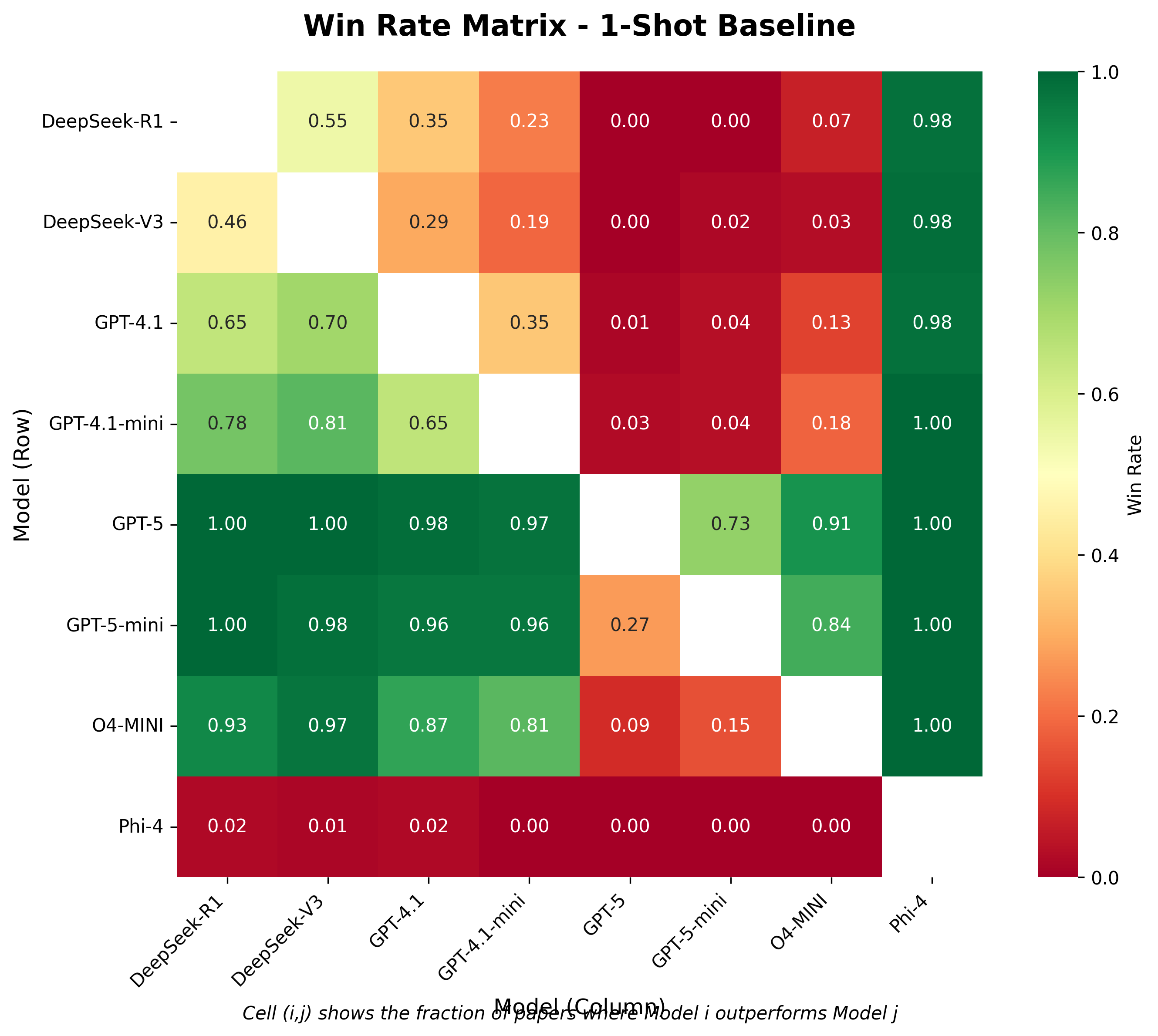} &
        \includegraphics[width=0.47\textwidth, trim={20 25 20 30}, clip]{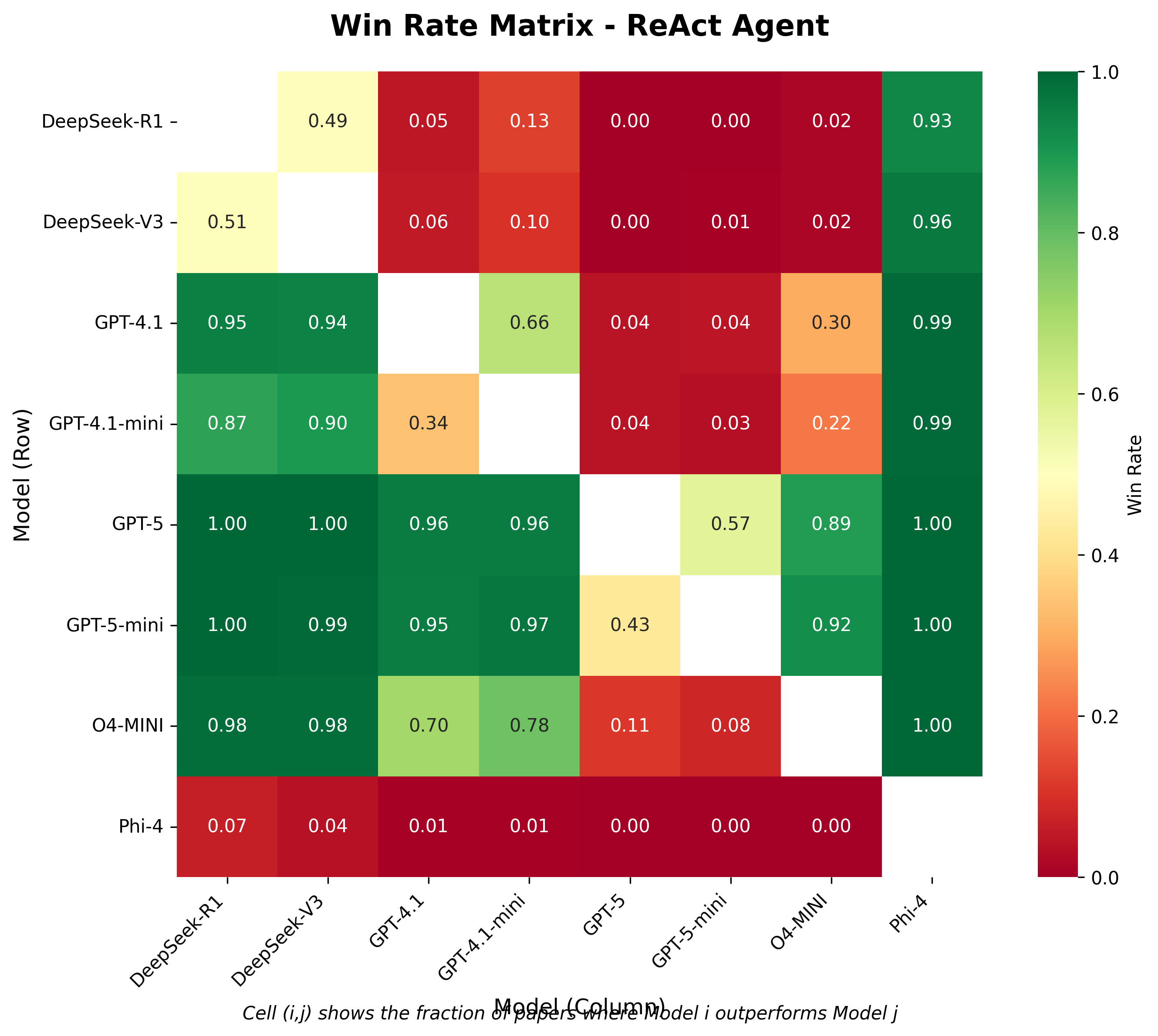} \\
        (c) 1-shot & (d) ReAct
    \end{tabular}
    \caption{Pairwise win rate matrices across prompting settings. Each cell $(i,j)$ shows the fraction of research ideas where model $i$ outperforms model $j$. The four panels correspond to (a) naïve prompting, (b) 0-shot, (c) 1-shot, and (d) ReAct settings. Across all settings, GPT-5 achieves the highest overall win rates.}
    \label{fig:win_rate_matrices}
\end{figure}

Figure~\ref{fig:win_rate_matrices} shows pairwise win rates under four prompting setups. GPT-5 consistently leads across naïve, 0-shot, 1-shot, and ReAct settings, though gains narrow in the ReAct case.

\section{Case Study: Knowledge Conflict in ReAct Agent}
\label{sec:case_study_knowledge_conflict}

Table~\ref{tab:react_knowledge_conflict} illustrates a case where the ReAct agent retrieves papers that conflict with its parametric knowledge, leading to less relevant citations compared to the baseline. While the agent cites specific technical papers on empirical welfare maximization, the baseline correctly identifies the foundational ``Prediction Policy Problems'' paper~\citep{kleinberg2015prediction} from its training data, demonstrating that retrieval can introduce noise when it does not align with the model's existing domain knowledge.

\begin{table}[htbp]
\centering
\caption{Why React Agent Does Not Outperform Baseline: Parametric vs Retrieved Knowledge Conflict. 
Green text indicates the citations that match the ground truth and are retrieved from the model’s parametric knowledge. 
This case shows that GPT-5 can correctly identify the key citation from its internal knowledge, but retrieval introduces noise when it conflicts with the model’s domain understanding, suggesting the need for further study to mitigate such effects.}
\label{tab:react_knowledge_conflict}
\small

\begin{tabular}{|p{0.95\textwidth}|}
\hline
\textbf{Ground Truth} \\
\hline
\textbf{Domain: Prediction-Policy Problems and Screening Frameworks}

$\bullet$ \textcolor{green}{Kleinberg, J., Ludwig, J., Mullainathan, S., \& Obermeyer, Z. (2015). "Prediction Policy Problems." American Economic Review, 105(5):491–495.}
-- Introduces the paradigm of separately trained predictive risk scoring systems for policy allocation problems, motivating the need to quantitatively compare prediction accuracy against resource constraints.

$\bullet$ Perdomo, J. C. (2024). "The Relative Value of Prediction in Algorithmic Decision Making." ICML 2024.
-- Defines the prediction-access ratio (PAR) in binary treatment settings; serves as the foundation for extending PAR to continuous welfare screening and developing theoretical bounds. \\
\hline
\end{tabular}

\vspace{0.3cm}

\begin{tabular}{|p{0.46\textwidth}|p{0.46\textwidth}|}
\hline
\textbf{0-shot (GPT-5)} & \textbf{React Agent (GPT-5)} \\
\hline
\textbf{Prediction policy and algorithmic decision-making} &
\textbf{Policy learning and empirical welfare maximization under constraints} \\

$\bullet$ \textcolor{green}{Kleinberg, Ludwig, Mullainathan, and Sunstein on prediction policy problems. Connects predictive accuracy to policy outcomes and clarifies when prediction improves decisions.} &
$\bullet$ Model Selection for Treatment Choice: Penalized Welfare Maximization (arXiv:1609.03167): Extends EWM to select among constrained policy classes with oracle regret guarantees; key for interpretable policy classes and complexity control. \\

$\bullet$ Obermeyer and Mullainathan on algorithmic bias in healthcare risk prediction. Highlights label/measurement biases that directly inform our decomposition and robustness checks. &
$\bullet$ Policy Learning with Observational Data (Athey \& Wager; arXiv:1702.02896): Doubly robust policy learning with optimal n\^{}(-1/2) regret for VC classes and accommodation of observational designs; central for estimating welfare-maximizing targeting rules from administrative data. \\

$\bullet$ Bertsimas and Kallus on prescriptive analytics. Informs policy optimization from predictive models under constraints. & \\
\hline
\end{tabular}
\end{table}

\section{Enhanced Context Experiment}\label{appen:enhanced_context_experiment}

To investigate whether providing curated related literature improves research plan quality, we design an experiment where models are given three key foundational papers selected by the Literature Recommendation Prompt (Table~\ref{tab:literature_recommendation}). Table~\ref{tab:baseline_related} shows the section-wise performance comparison with and without curated literature across all models. The results demonstrate that providing relevant foundational papers leads to modest improvements in overall performance for most models, with the largest gains observed in the Introduction and Literature sections.

\begin{table}[h]
\centering
\caption{Literature Recommendation Prompt}
\label{tab:literature_recommendation}
\tiny
\fbox{\begin{tabular}{p{0.93\textwidth}}
\textbf{Literature Recommendation Prompt} \\
\\
You are a research literature recommendation system. Based on the provided research idea, introduction, and references, identify exactly 3 most relevant key papers that researchers should read to better understand and build upon this work. \\
\\
\#\# RESEARCH IDEA: \\
\{\{RESEARCH\_IDEA\}\} \\
\\
\#\# INTRODUCTION: \\
\{\{INTRODUCTION\}\} \\
\\
\#\# REFERENCES: \\
\{\{REFERENCES\}\} \\
\\
\#\# TASK: \\
Instructions: \\
1. Carefully analyze the research idea to understand its core contributions, methods, and objectives \\
2. Read the introduction to identify the key technical foundations and prior work this research builds upon \\
3. Select exactly 3 papers that serve as direct foundational works by being: \\
~~~~- Core methodological predecessors that this work extends or improves \\
~~~~- Papers introducing the key techniques, algorithms, or frameworks being built upon \\
~~~~- Essential baseline methods or datasets that this work directly compares against or uses \\
\\
~~~Note: Focus on papers that are specifically related to understanding this specific research idea, avoid general background papers. \\
\\
\#\# OUTPUT FORMAT: \\
Return your response as a JSON array with exactly 3 papers. Each paper should have: \\
- "title": The exact title of the paper \\
- "arxiv\_link": The arXiv ID if available (format: "XXXX.XXXXX" e.g., "1706.03762"), or null if not on arXiv \\
- "relevance": A single sentence explaining why this paper is relevant to the current research (max 30 words) \\
\\
Example format: \\
\\
\texttt{[} \\
\texttt{~~\{} \\
\texttt{~~~~"title": "Attention Is All You Need",} \\
\texttt{~~~~"arxiv\_link": "1706.03762",} \\
\texttt{~~~~"relevance": "Introduces the transformer architecture which forms the foundation for the proposed modifications in attention mechanisms."} \\
\texttt{~~\},} \\
\texttt{~~\{} \\
\texttt{~~~~"title": "BERT: Pre-training of Deep Bidirectional Transformers for Language Understanding",} \\
\texttt{~~~~"arxiv\_link": "1810.04805",} \\
\texttt{~~~~"relevance": "Demonstrates successful pre-training strategies that are extended in this work for domain-specific applications."} \\
\texttt{~~\}} \\
\texttt{]} \\
\\
Return only the JSON array, no additional text or explanation. \\
\end{tabular}}
\end{table}

\begin{table}[t]
\centering
\footnotesize
\setlength{\tabcolsep}{4pt}
\caption{Section-wise performance comparison with and without curated literature. Values are shown as baseline\%/with-related-papers\%, with percentage point deltas in parentheses.}
\label{tab:baseline_related}
\begin{tabular}{lcccccc}
\toprule
Model & Intro & Lit & Methods & Exp Design & Resources \& Ethics & Overall \\
\midrule
DeepSeek-V3 & 39.3/41.2 & 29.7/31.8 & 37.8/39.1 & 32.8/33.4 & 44.2/45.9 & 32.7/35.1 \\
& (+1.9) & (+2.1) & (+1.3) & (+0.6) & (+1.7) & (+2.3) \\
DeepSeek-R1 & 39.8/42.3 & 31.6/31.8 & 42.8/41.2 & 35.7/36.1 & 49.2/50.9 & 37.2/38.5 \\
& (+2.5) & (+0.2) & (-1.6) & (+0.3) & (+1.7) & (+1.3) \\
GPT-4.1-mini & 42.5/43.9 & 25.1/30.9 & 42.3/40.3 & 37.4/37.3 & 48.1/45.7 & 35.3/37.5 \\
& (+1.4) & (+5.8) & (-2.0) & (-0.1) & (-2.4) & (+2.2) \\
GPT-4.1 & 40.5/43.7 & 25.3/31.1 & 44.1/43.0 & 36.6/39.9 & 50.1/49.1 & 35.2/38.7 \\
& (+3.2) & (+5.9) & (-1.1) & (+3.3) & (-1.1) & (+3.5) \\
o4-mini & 50.6/51.1 & 30.0/32.0 & 51.7/51.0 & 46.0/45.1 & 58.1/57.7 & 45.1/46.1 \\
& (+0.5) & (+2.0) & (-0.7) & (-0.9) & (-0.4) & (+1.0) \\
GPT-5-mini & 56.1/58.9 & 38.6/40.6 & 64.7/65.9 & 57.7/58.7 & 66.9/69.3 & 56.0/58.0 \\
& (+2.8) & (+2.0) & (+1.2) & (+1.0) & (+2.4) & (+2.1) \\
GPT-5 & 59.4/60.6 & 46.0/42.1 & 71.7/71.4 & 62.4/60.6 & 71.9/73.8 & 61.3/61.5 \\
& (+1.2) & (-3.9) & (-0.2) & (-1.8) & (+2.0) & (+0.2) \\
\bottomrule
\end{tabular}
\end{table}

\section{Newer Model and Agentic Harness Evaluation}
\label{appen:icml2026_codex}

To test whether newer models and agentic coding harnesses close the gap on \emph{Idea2Plan}, we construct an additional contamination-free benchmark from ICML 2026 papers. Since GPT-5.5 has a newer knowledge cutoff of December 1, 2025, we sample 100 ICML 2026 spotlight and oral papers that were not publicly available before this date. We apply the same automatic benchmark construction pipeline used in the main paper to generate \emph{Idea2Plan-ICML2026}.

We evaluate GPT-5.5 both as a standalone model and inside the Codex harness. For GPT-5.5 with Codex, we provide access to the Asta Scientific Corpus MCP tool\footnote{\url{https://allenai.org/asta/resources/mcp}}, which supports searching and reading academic papers. To avoid contamination, we set the publication-date cutoff for the Asta MCP tool to December 1, 2025. On average, Codex makes 20.7 Asta MCP calls per paper.

\begin{table}[H]
\centering
\footnotesize
\caption{Performance on \emph{Idea2Plan-ICML2026}. Scores are section-wise Planning Scores in percentages.}
\label{tab:icml2026_codex}
\begin{tabular}{lrrrrrr}
\toprule
Model & Overall & Intro & Key Lit. & Methods & Exp. & Res/Eth \\
\midrule
GPT-5.5 & 62.0 & 84.9 & 17.5 & 81.5 & 67.3 & 59.0 \\
GPT-5.5 + Codex & 75.1 & 92.5 & 40.5 & 91.8 & 80.5 & 70.0 \\
Upper bound & 85.8 & 95.9 & 89.4 & 98.7 & 93.5 & 51.7 \\
\bottomrule
\end{tabular}
\end{table}

Table~\ref{tab:icml2026_codex} shows that \emph{Idea2Plan} remains unsolved even with newer models and stronger agentic harnesses. GPT-5.5 with Codex improves substantially over standalone GPT-5.5, especially in the Key Literatures section, where it gains 23.0 points. However, it still trails the upper bound by 48.9 points in that section. This suggests that agentic paper-search tools can improve research planning by identifying more relevant prior work, but current systems still struggle to recover the key literature needed for high-quality research plans. In Methods and Experimental Design, GPT-5.5 with Codex is closer to the upper bound, indicating that the primary remaining bottleneck is grounding the plan in the right scientific context.

\section{SFT Experiment}\label{appen:sft_experiment}

Supervised fine-tuning (SFT) has been a common strategy to enhance LLMs for specialized domains~\citep{finetune_llm}. Recent studies show that SFT on scientific data can improve instruction-following ability of LLMs~\citep{SciInstruct, SciRIFF, SciLitLLM}. However,~\citet{fine_tune_hallucination} show that fine-tuning on factual knowledge absent from an LLM’s pretraining corpus can increase the tendency to hallucinate.
For the \emph{Idea2Plan} task, one natural source of high-quality supervision is the corpus of published research papers. We can extract research ideas with their corresponding research plans from these papers, which are appealing as potential training data. 
To examine whether fine-tuning on research papers can improve LLM's performance on \emph{Idea2Plan}, we fine-tune GPT-4.1-mini and GPT-4.1 using research papers from ICML 2024. From this corpus, we extract 2{,}000 idea--plan pairs, which serve as our training set. We then fine-tune GPT-4.1-mini and GPT-4.1 on this data using the Azure finetuning API\footnote{\url{https://learn.microsoft.com/en-us/azure/ai-foundry/concepts/fine-tuning-overview}} and compare their performance against their base versions in the 1-shot setting. 

Table~\ref{tab:sft_comparison} summarizes the results. SFT leads to a notable drop in overall performance. The largest degradation occurs in the ethics and resources dimension, where research papers often omit practical details such as GPU hours, causing the fine-tuned models to omit this part. Even after excluding this section from the aggregate scores, the SFT models underperform their baselines. This suggests that direct training on research plans does not reliably improve research planning ability. We also observe increased hallucination, particularly in the related-literature section. A possible reason is that extracted research plans contain factual details (\eg specific papers or datasets) absent from the models’ pretraining data, which the models interpret during fine-tuning as a signal to hallucinate.

    \paragraph{Future directions.} 
    Our current fine-tuning dataset provides limited coverage of the ethics and resources considerations in the research plans. To address this gap, future work could augment the fine-tuning data with synthetic examples of these aspects.
    It is also worthwhile to explore RL-based training (\eg PPO~\citep{PPO}) in an agentic loop where the model drafts plans, receives rubric feedback, and optimizes research plan generation quality.

\begin{table}[t]
\centering
\footnotesize
\setlength{\tabcolsep}{4pt}
\caption{SFT model comparison on 1-shot vanilla baseline agents. Scores are averaged across sections and reported in percentages. Rows labeled $ \Delta = \text{SFT} - \text{Base} $ show percentage-point differences between SFT and Base models.}
\label{tab:sft_comparison}
\begin{tabular}{lccccccc}
\toprule
Model & Intro & Lit & Method & Exp & Res/Eth & Overall & Overall (w/o Res/Eth) \\
\midrule
GPT-4.1-mini (Base) & 48.2\% & 26.5\% & 40.4\% & 33.8\% & 65.3\% & 42.8\% & 37.2\% \\
GPT-4.1-mini (SFT)  & 42.4\% & 22.9\% & 41.5\% & 27.2\% & 18.9\% & 30.6\% & 33.5\% \\
$ \Delta $ (SFT–Base) & -5.7\% & -3.6\% & +1.1\% & -6.6\% & -46.4\% & -12.2\% & -3.7\% \\
\midrule
GPT-4.1 (Base)      & 45.2\% & 24.0\% & 38.1\% & 31.7\% & 65.1\% & 40.8\% & 34.8\% \\
GPT-4.1 (SFT)       & 44.5\% & 22.1\% & 41.7\% & 27.9\% & 21.9\% & 31.6\% & 34.1\% \\
$ \Delta $ (SFT–Base) & -0.7\% & -1.9\% & +3.6\% & -3.8\% & -43.2\% & -9.2\% & -0.7\% \\
\bottomrule
\end{tabular}
\end{table}

\section{Nature Mental Health Dataset}
\label{appen:nmh_dataset}

To test the generalizability of our pipeline for generating research plans beyond AI research, we construct an additional dataset from Nature Mental Health (NMH). NMH is a peer-reviewed journal with a 2024 impact factor of 8.7, focusing on mental health and mental health disorders. This represents a distinct domain from AI research.

\subsection{Data Collection and Template Adaptation}

We sample the most recent articles from NMH.\footnote{\url{https://www.nature.com/natmentalhealth/articles?type=article}. Retrieved on 2025-12-14.} All sampled papers were published after February 2025. We further exclude articles that had a publicly available version on Semantic Scholar\footnote{\url{https://www.semanticscholar.org/}} before 2025, resulting in a final set of 46 articles. 
This ensures they are free from potential training data contamination for the LLMs we evaluate. 

According to the formatting conventions of NMH, articles typically follow a structure consisting of an opening Introduction, followed by Results, Discussion, and Methods sections. To align with this structure while maintaining consistency with our evaluation framework, we adapt our research plan template to four sections: Introduction, Key Literatures, Method, and Initial Experimental Design. This adaptation preserves the core components necessary for research planning evaluation while respecting domain-specific conventions.

We apply the same automated extraction pipeline used for the ICML dataset, employing o4-mini (reasoning=high) to extract research ideas, reference plans, and grading rubrics from each paper.

\subsection{Prompt Templates and Baselines}

We provide the prompt templates used for both baselines on the NMH dataset:
\begin{itemize}
\item \textit{Naive Baseline Template:} Table~\ref{tab:nmh_naive_template} presents the basic template with section headers only, used for the naive baseline.
\item \textit{1-Shot Baseline Template:} Tables~\ref{tab:nmh_1shot_template_part1} and~\ref{tab:nmh_1shot_template_part2} provide the detailed template with descriptions and a full example, used for the 1-shot baseline.
\end{itemize}

We evaluate over the naive baseline and the 1-shot baseline. For the 1-shot baseline, we use the research plan from one NMH article outside of the test set as the demonstration example; this example is manually curated to ensure quality.

\subsection{Experimental Results}
\label{appen:nmh_results}

Table~\ref{tab:nmh_results} presents the section-wise planning scores on the NMH dataset. We evaluate GPT-5, o4-mini, and DeepSeek-R1 using both naive and 1-shot baselines across three independent runs per paper. We observe consistent patterns with our ICML results. GPT-5 achieves the highest scores, followed by o4-mini and DeepSeek-R1. The literature section remains the most challenging across all models, while all three models show strong performance on method sections.

\begin{table}[h]
\centering
\footnotesize
\caption{Section-wise Planning Scores (\%) on the Nature Mental Health dataset with Mean aggregation. Each value represents the mean accuracy across all papers. Bold numbers indicate the highest score within each section across all baselines. We also include an \emph{upper-bound} in which o4-mini is given the original paper to generate a plan, approximating the highest achievable performance.}
\label{tab:nmh_results}
\begin{tabular}{l:rrrr:r:rrrr:r}
\toprule
\multirow{2}{*}{Model} & \multicolumn{5}{c:}{Naive} & \multicolumn{5}{c}{1-Shot} \\
\cmidrule(lr){2-6} \cmidrule(lr){7-11}
 & Intro & Lit & Met & Exp & Avg & Intro & Lit & Met & Exp & Avg \\
\midrule
GPT-5 & \textbf{75.0} & \textbf{45.4} & \textbf{87.8} & \textbf{75.7} & \textbf{71.0} & \textbf{75.0} & \textbf{48.0} & \textbf{85.8} & \textbf{73.8} & \textbf{70.7} \\
o4-mini & 70.6 & 38.5 & 78.6 & 68.4 & 64.0 & 68.9 & 37.8 & 77.2 & 63.9 & 61.9 \\
DeepSeek-R1 & 69.5 & 34.6 & 66.3 & 58.9 & 57.3 & 69.9 & 35.7 & 65.4 & 58.2 & 57.3 \\
\midrule
Upper-bound & \multicolumn{10}{c}{98.4} \\
\bottomrule
\end{tabular}
\end{table}

\begin{table}[h]
\centering
\caption{Nature Mental Health Naive Baseline Template}
\label{tab:nmh_naive_template}
\tiny
\fbox{\begin{tabular}{p{0.93\textwidth}}
\textbf{Naive Baseline Template for Nature Mental Health Dataset} \\
\\
\#\# 1. Introduction \\
\\
\#\#\# 1.1 Background \\
\\
\#\#\# 1.2 Primary Objectives \\
\\
\#\#\# 1.3 Research Questions \\
\\
\#\# 2. Key Literatures \\
\\
\#\# 3. Methods \\
\\
\#\# 4. Initial Experimental Design \\
\end{tabular}}
\end{table}

\begin{table}[h]
\centering
\caption{Nature Mental Health 1-Shot Baseline Template, Part 1}
\label{tab:nmh_1shot_template_part1}
\tiny
\fbox{\begin{tabular}{p{0.93\textwidth}}
\textbf{1-Shot Baseline Template for Nature Mental Health Dataset} \\
\\
\#\# 1. Introduction \\
\\
\#\#\# 1.1 Background \\
Describe the current limitations or challenges in the field and why they matter now. This should be one paragraph. \\
\\
\#\#\# 1.2 Primary Objectives \\
List specific, measurable goals that define project success. These should align with the contributions typically outlined in a paper's introduction. \\
\\
\#\#\# 1.3 Research Questions \\
State the core research questions for this research project. \\
\\
\#\# 2. Key Literatures \\
Identify key related works and relevant domains that inform your research. Begin by listing a few key domains, and for each cited work, briefly explain why it is important to your current research plan. \\
\\
\#\# 3. Methods \\
Describe the methodology, e.g., study design, data collection procedures, measurement instruments, statistical analysis approaches, etc. \\
\\
\#\# 4. Initial Experimental Design \\
Describe the experimental design, e.g., sample selection, data sources, outcome measures, analysis plan, etc. \\
\\
--- \\
\\
\#\# Examples \\
\\
\textbf{IMPORTANT NOTE}: The following examples are provided to illustrate the expected format and level of detail for each section. These examples are from a specific research plan. However, you should NOT be restricted to the specific settings, methods, datasets, or approaches mentioned in these examples. Adapt all content to fit the specific research idea you are working on. Use your domain knowledge and the information you gather to create a plan that is most appropriate for your particular research context. \\
\end{tabular}}
\end{table}

\begin{table}[h]
\centering
\caption{Nature Mental Health 1-Shot Baseline Template, Part 2: Introduction \& Key Literatures}
\label{tab:nmh_1shot_template_part2}
\tiny
\fbox{\begin{tabular}{p{0.93\textwidth}}
\#\# 1. Introduction \\
\\
\#\#\# 1.1 Background \\
\\
- Interest in human flourishing has expanded across psychology, economics, business, education, medicine, public health and public policy, but much of the research remains rooted in Western contexts. \\
- Existing cross-national studies (World Values Survey; Gallup World Poll/World Happiness Report) are largely cross-sectional and focus on life evaluation or happiness rather than a broad, multidimensional conception of flourishing. \\
- There is a need for longitudinal, panel data collected from the same individuals across diverse cultural and geographic settings, with a broader range of well-being assessments, to understand both universal and culturally specific determinants of flourishing. \\
\\
\#\#\# 1.2 Primary Objectives \\
\\
1. To map the global distribution of a composite flourishing index across 22 geographically and culturally diverse countries. \\
2. To examine associations between composite flourishing and key demographic characteristics (age, gender, marital status, employment status, education level, religious service attendance, and immigration status), both pooled and within countries. \\
3. To evaluate how retrospective assessments of childhood experiences (parental relationships; parental marital status; family financial status in childhood; abuse; feeling like an outsider; self-rated childhood health; childhood religious service attendance; immigration status) relate to adult flourishing. \\
4. To identify which demographic and childhood predictors of flourishing are consistent across all countries (universal) and which vary by country (culturally specific). \\
5. To establish a five-year longitudinal panel (five annual waves) enabling future analysis of within-person changes and causal inferences regarding flourishing. \\
\\
\#\#\# 1.3 Research Questions \\
\\
- RQ1: How does composite flourishing vary in mean level and distribution across 22 countries? \\
- RQ2: What are the associations between demographic characteristics and composite flourishing, both overall and within each country? \\
- RQ3: What are the associations between specific retrospective childhood experiences and adult composite flourishing, both overall and by country? \\
- RQ4: Which demographic and childhood experience associations with flourishing are universal and which are culturally specific? \\
\\
\#\# 2. Key Literatures \\
\\
- VanderWeele TJ (2017). "On the promotion of human flourishing." \textit{Proc. Natl. Acad. Sci. USA} 114:8148–8156. \\
\quad - Provided a working multidimensional definition of flourishing and introduced the composite flourishing index of 12 self-report indicators across six domains, which underpins the present study. \\
- Huppert FA \& So TT (2013). "Flourishing across Europe: application of a new conceptual framework for defining well-being." \textit{Soc. Indic. Res.} 110:837–861. \\
\quad - Offered a conceptual framework for multidimensional flourishing domains, informing the domain structure (health, happiness, meaning, character, relationships, material well-being) used in measurement. \\
- Keyes CL (2002). "The mental health continuum: from languishing to flourishing in life." \textit{J. Health Soc. Behav.} 43:207–222. \\
\quad - Introduced the mental health continuum model, distinguishing between states of languishing, moderate mental health, and flourishing, providing theoretical grounding for well-being as more than the absence of ill-being. \\
- World Values Survey (ongoing). \\
\quad - A large cross-national survey capturing values and well-being measures; serves as a key baseline for international comparisons but lacks longitudinal panel follow-up of the same individuals. \\
- Helliwell JF, Layard R, Sachs JD, De Neve JE (2021). "World Happiness Report 2021." Sustainable Development Solutions Network. \\
\quad - Offers annual nation-level rankings of life evaluation and happiness, providing comparative benchmarks but limited in breadth of flourishing domains and in panel design. \\
- World Health Organization (1948). "Preamble to the Constitution of the World Health Organization." \\
\quad - Defined health as "a state of complete physical, mental, and social well-being," informing the inclusion of physical, emotional, cognitive, volitional, social and material domains in flourishing measurement. \\
\end{tabular}}
\end{table}

\begin{table}[h]
\centering
\caption{Nature Mental Health 1-Shot Baseline Template, Part 3: Methods \& Experimental Design}
\label{tab:nmh_1shot_template_part3}
\tiny
\fbox{\begin{tabular}{p{0.93\textwidth}}
\#\# 3. Methods \\
\\
\textbf{Study Design:} \\
- Five-year longitudinal panel study (Wave 1 in 2023, followed by annual Waves 2–5 through 2027). \\
- Sample: 202,898 adults ($\geq$18 years) drawn via nationally representative sampling in 22 countries spanning all six populated continents. Country sample sizes ranged from 1,473 (Turkey) to 38,312 (United States). \\
- Country Selection Criteria: maximize global population coverage; ensure geographic, cultural, religious diversity; leverage existing data collection infrastructure. \\
\\
\textbf{Data Collection:} \\
- Conducted by Gallup Inc. under IRB approval (Gallup; Baylor University), using web and other self-administered modes. Participants received modest compensation (US \$3–6). \\
- Panel Retention: The same cohort is surveyed annually over five years to permit within-person longitudinal analysis. \\
- Public Access: Wave 1 data available through the Center for Open Science after pre-registration; full open release planned in 2026. \\
\\
\textbf{Measures:} \\
- Outcome – Composite flourishing index: mean of 12 self-report items (0–10 scale) covering six domains (happiness, health, meaning, character, relationships, financial security). \\
- Demographics – Age group (18–24, 25–29, 30–39, 40–49, 50–59, 60–69, 70–79, $\geq$80), gender (male, female, other), marital status, employment status, education ($\leq$8 yrs, 9–15 yrs, $\geq$16 yrs), religious service attendance (>1 wkly, 1 wkly, 1–3 mo, a few/yr, never), immigration status (born in country vs born elsewhere). \\
- Childhood Predictors – Retrospective self-reports: relationship quality with mother/father (good vs bad), parental marital status (married, divorced, never married, one/both deceased), family financial status at $\sim$age 12 (lived comfortably, got by, found it difficult, found it very difficult), history of physical/sexual abuse (yes/no), feeling like an outsider (yes/no), self-rated health in childhood (excellent to poor), religious service attendance at age 12 ($\geq$1 wkly, 1–3 mo, <1 mo, never), childhood immigration status. \\
\\
\textbf{Statistical Analysis:} \\
- Descriptive statistics: Weighted means and proportions for all variables by country. \\
- Primary analyses: Random-effects meta-analysis pooling country-specific associations between demographics/childhood predictors and flourishing outcomes. Report means, 95\% CIs, heterogeneity metrics ($\tau$, I$^2$), and Bonferroni-corrected P-values. \\
- Missing data: Multiple imputation by chained equations. \\
- Sensitivity analyses: E-values to assess unmeasured confounding; alternative model specifications. \\
- Software: R (metafor, mice) for analyses; pre-registration and code available via OSF. \\
\\
\#\# 4. Initial Experimental Design \\
\\
- Descriptive mapping of composite flourishing across 22 countries, including mean levels and distributional patterns. \\
- Estimation of demographic associations with composite flourishing within each country and pooled across countries. \\
- Estimation of associations between retrospective childhood experiences and adult composite flourishing within each country and pooled across countries. \\
- Quantification of cross-national heterogeneity and identification of universal versus culturally specific determinants. \\
\end{tabular}}
\end{table}

\section{ReAct Agent Retrieval Challenges}
\label{appen:react_retrieval}

While the ReAct agent has access to arXiv search and read tools, its performance is significantly affected by search engine retrieval quality. Analysis of tool-call history reveals that search engines often prioritize recent papers over foundational earlier work, even when queries explicitly mention the foundational paper's authors and title. For example, Table~\ref{tab:react_retrieval_challenge} shows a case where the key reference is \textit{Prediction Policy Problems} (2015), yet all retrieved papers are from 2016 onwards, suggesting that temporal bias in retrieval can mislead the agent by providing recent work that may not capture the original concepts needed for the research plan.

\begin{table}[h]
\centering
\small
\caption{Example of search engine temporal bias affecting ReAct agent retrieval. Despite the search query explicitly mentioning the 2015 foundational paper by name and authors, the Bing search engine returns only papers from 2016 onwards.}
\label{tab:react_retrieval_challenge}
\begin{tabular}{p{0.95\textwidth}}
\toprule
\textbf{Ground Truth Reference (from rubric):} \\
Kleinberg, J., Ludwig, J., Mullainathan, S., \& Obermeyer, Z. (2015). "Prediction Policy Problems." \textit{American Economic Review}, 105(5):491–495. \\
\midrule
\textbf{Agent's Search Queries (via Bing Search):} \\
1. "prediction policy problems welfare government programs machine learning Kleinberg Mullainathan Obermeyer arXiv" \\
2. "Prediction Policy Problems Kleinberg Ludwig Mullainathan Obermeyer arXiv" \\
\midrule
\textbf{Papers Returned by Bing Search (all published after 2015):} \\
• 2016: "Inherent Trade-Offs in the Fair Determination of Risk Scores" \\
• 2018: "Direct Uncertainty Prediction for Medical Second Opinions" \\
• 2018: "Simplicity Creates Inequity" \\
• 2019: "The Algorithmic Automation Problem" \\
• 2019: "Machine learning in policy evaluation" \\
• 2019: "Discrimination in the Age of Algorithms" \\
• 2021: "Algorithmic Monoculture and Social Welfare" \\
• 2023: "Policy Learning with Distributional Welfare" \\
• 2023: "Transparency challenges in policy evaluation" \\
• 2025: "Optimal Policy Adaptation under Covariate Shift" \\
\bottomrule
\end{tabular}
\end{table}

Possible mitigations include configuring retrieval systems to return papers across broader time ranges, incorporating citation-based retrieval to trace backwards from recent papers to foundational work, or using hybrid approaches that combine search with citation graph traversal. However, these strategies require API-level control over search engines or custom retrieval systems, presenting practical challenges.

\begin{table}[h]
\centering
\caption{Evaluation rubric for Jailbreak-Tax paper~\citep{jailbreak_tax}, Part 1.}
\label{tab:rubric_jailbreak_tax_part1}
\vspace{0.2cm}
\tiny
\textbf{Evaluation Rubric Part 1 (JSON Format)}
\begin{lstlisting}[basicstyle=\tiny\ttfamily, breaklines=true]
{
  "sections": {
    "Introduction": {
      "subsections": {
        "Background": {
          "questions": [
            {
              "question": "Does the plan state that adversarial jailbreak attacks can bypass LLM safety guardrails or alignment protections?"
            },
            {
              "question": "Does the plan note that existing evaluations focus on bypass success and neglect post-jailbreak capability retention?"
            },
            {
              "question": "Does the plan explain that assessing the utility of harmful outputs requires domain expertise?"
            },
            {
              "question": "Does the plan highlight the lack of an unaligned baseline as a barrier to quantifying performance degradation?"
            },
            {
              "question": "Does the plan identify the need for a more rigorous evaluation framework for jailbreak attacks?"
            }
          ]
        },
        "Primary Objectives": {
          "questions": [
            {
              "question": "Does the plan propose designing an evaluation framework that measures both jailbreak success rate and the utility of elicited outputs?"
            },
            {
              "question": "Does the plan include constructing benchmark suites with tasks that have objectively verifiable ground-truth answers?"
            },
            {
              "question": "Does the plan include applying and comparing multiple representative jailbreak attacks across these benchmarks?"
            },
            {
              "question": "Does the plan include comparing jailbreak performance across different alignment methods?"
            },
            {
              "question": "Does the plan include quantifying the relative performance drop (jailbreak tax) compared to an unaligned baseline?"
            },
            {
              "question": "Does the plan aim to release benchmark suites and evaluation code to the community?"
            }
          ]
        },
        "Research Questions": {
          "questions": [
            {
              "question": "Does the plan ask whether different jailbreak attacks incur a measurable performance drop (jailbreak tax) across tasks?"
            },
            {
              "question": "Does the plan ask whether the magnitude of the performance drop correlates with the jailbreak success rate?"
            },
            {
              "question": "Does the plan ask whether model size influences the severity of the jailbreak tax?"
            },
            {
              "question": "Does the plan ask whether the jailbreak tax is consistent across alignment methods?"
            },
            {
              "question": "Does the plan ask whether task difficulty affects the magnitude of the jailbreak tax?"
            }
          ]
        }
      }
    }
  }
}
\end{lstlisting}
\end{table}

\begin{table}[h]
\centering
\caption{Evaluation rubric for Jailbreak-Tax paper~\citep{jailbreak_tax}, Part 2.}
\label{tab:rubric_jailbreak_tax_part2}
\vspace{0.2cm}
\tiny
\textbf{Evaluation Rubric Part 2 (JSON Format)}
\begin{lstlisting}[basicstyle=\tiny\ttfamily, breaklines=true]
{
  "sections": {
    "Key Literatures": {
      "subsections": {
        "Jailbreak Attack Methodologies": {
          "questions": [
            {
              "question": "Does the plan cite the paper (Jailbroken: How does LLM safety training fail?) or similar work on manual prompt-engineering jailbreaking techniques?"
            },
            {
              "question": "Does the plan cite the paper (Universal and transferable adversarial attacks on aligned language models) or similar work on optimization-based adversarial suffix attacks?"
            },
            {
              "question": "Does the plan cite the paper (AutoDAN: Generating stealthy jailbreak prompts on aligned large language models) or similar work on genetic-algorithm-based prompt generation?"
            },
            {
              "question": "Does the plan cite the paper (Jailbreaking black box large language models in twenty queries) or similar work on iterative LLM-based prompt rewriting attacks?"
            },
            {
              "question": "Does the plan cite the paper (Tree of attacks: Jailbreaking black-box LLMs automatically) or similar work on tree-of-thought or hierarchical attack expansions?"
            },
            {
              "question": "Does the plan cite the paper (Many-shot jailbreaking) or similar work on in-context demonstration-based jailbreak attacks?"
            },
            {
              "question": "Does the plan cite the paper (Multilingual jailbreak challenges in large language models) or similar work on translation-based or multilingual jailbreak techniques?"
            }
          ]
        },
        "Benchmark Datasets for Objective Utility": {
          "questions": [
            {
              "question": "Does the plan cite the paper (The WMDP benchmark: Measuring and reducing malicious use with unlearning) or similar work on bio-security multiple-choice benchmarks?"
            },
            {
              "question": "Does the plan cite the paper (Training verifiers to solve math word problems) or similar work on grade-school math reasoning benchmarks?"
            },
            {
              "question": "Does the plan cite the paper (Measuring massive multitask language understanding) or similar work on competition-level math or multitask reasoning benchmarks (e.g., MATH, MMLU)?"
            }
          ]
        },
        "Alignment Tax and Prior Evaluations": {
          "questions": [
            {
              "question": "Does the plan cite the paper (Current work in AI alignment) or similar work defining the concept of an alignment tax?"
            },
            {
              "question": "Does the plan cite the paper (A StrongReject for empty jailbreaks) or similar work on performance degradation of jailbreak attacks?"
            },
            {
              "question": "Does the plan cite the paper (AgentHarm: A benchmark for measuring harmfulness of LLM agents) or similar work contrasting objective benchmarks with LLM-based evaluation of harmfulness?"
            }
          ]
        }
      }
    }
  }
}
\end{lstlisting}
\end{table}

\begin{table}[h]
\centering
\caption{Evaluation rubric for Jailbreak-Tax paper~\citep{jailbreak_tax}, Part 3.}
\label{tab:rubric_jailbreak_tax_part3}
\vspace{0.2cm}
\tiny
\textbf{Evaluation Rubric Part 3 (JSON Format)}
\begin{lstlisting}[basicstyle=\tiny\ttfamily, breaklines=true]
{
  "sections": {
    "Methods": {
      "subsections": {
        "Pseudo-Alignment Techniques": {
          "questions": [
            {
              "question": "Does the plan describe a system-prompt alignment method that simulates model refusals via instructional constraints?"
            },
            {
              "question": "Does the plan describe a supervised fine-tuning approach on (prompt, refusal) pairs to induce model alignment?"
            },
            {
              "question": "Does the plan include a data-driven or task rewording-based pseudo-alignment technique (e.g., transforming benign tasks to harmful contexts)?"
            }
          ]
        },
        "Jailbreak Attacks": {
          "questions": [
            {
              "question": "Does the plan include a prompt-based jailbreak attack that appends or modifies instructions to bypass alignment?"
            },
            {
              "question": "Does the plan include a fine-tuning-based jailbreaking approach to reverse the model's refusal alignment?"
            },
            {
              "question": "Does the plan include demonstration-based (many-shot) jailbreak attacks using in-context examples?"
            },
            {
              "question": "Does the plan include optimization-based adversarial attacks (e.g., greedy coordinate descent) to craft adversarial suffixes?"
            },
            {
              "question": "Does the plan include genetic algorithm or evolutionary strategies for generating stealthy jailbreak prompts?"
            },
            {
              "question": "Does the plan include translation-based or multilingual jailbreak attacks?"
            },
            {
              "question": "Does the plan include iterative prompt rewriting attacks that use LLM-based rewriting (with or without tree-of-thought reasoning)?"
            }
          ]
        },
        "Evaluation Metrics": {
          "questions": [
            {
              "question": "Does the plan define a metric for jailbreak success rate that measures the proportion of non-refusal outputs?"
            },
            {
              "question": "Does the plan define a metric for post-jailbreak utility, such as conditional task accuracy after a successful jailbreak?"
            },
            {
              "question": "Does the plan define a baseline utility metric for measuring the unaligned model's task performance?"
            },
            {
              "question": "Does the plan define a metric for quantifying the relative performance drop (jailbreak tax) between baseline and jailbroken outputs?"
            }
          ]
        }
      }
    }
  }
}
\end{lstlisting}
\end{table}

\begin{table}[h]
\centering
\caption{Evaluation rubric for Jailbreak-Tax paper~\citep{jailbreak_tax}, Part 4.}
\label{tab:rubric_jailbreak_tax_part4}
\vspace{0.2cm}
\tiny
\textbf{Evaluation Rubric Part 4 (JSON Format)}
\begin{lstlisting}[basicstyle=\tiny\ttfamily, breaklines=true]
{
  "sections": {
    "Initial Experimental Design": {
      "questions": [
        {
          "question": "Does the plan propose experiments that evaluate both jailbreak success rate and post-jailbreak utility on objective benchmarks?"
        },
        {
          "question": "Does the plan include experiments using tasks with verifiable ground-truth answers (e.g., biology multiple-choice, grade-school math, competition-level math)?"
        },
        {
          "question": "Does the plan include experiments comparing results across multiple model sizes (e.g., small, medium, large LLMs)?"
        },
        {
          "question": "Does the plan include experiments comparing different alignment methods (e.g., system-prompt, supervised fine-tuning, task rewording)?"
        },
        {
          "question": "Does the plan include experiments applying multiple jailbreak attacks across the selected benchmarks?"
        },
        {
          "question": "Does the plan include analysis of the relationship between task difficulty and the magnitude of the jailbreak tax?"
        },
        {
          "question": "Does the plan include control experiments evaluating aligned models on neutral or unrelated tasks (e.g., subsets of MMLU or other benchmarks) to verify unintended degradation?"
        }
      ]
    },
    "Resources, Compliance, and Ethical Considerations": {
      "subsections": {
        "Resource Requirements": {
          "questions": [
            {
              "question": "Does the plan list required model resources, such as open-weight and closed-weight LLMs?"
            },
            {
              "question": "Does the plan acknowledge the need for computational infrastructure for fine-tuning and evaluation (e.g., GPU hardware)?"
            },
            {
              "question": "Does the plan acknowledge the need for API access for data transformation or evaluation steps?"
            }
          ]
        },
        "Ethical and Compliance Considerations": {
          "questions": [
            {
              "question": "Does the plan address the potential for adversaries to misuse the evaluation framework?"
            },
            {
              "question": "Does the plan mention verifying data licenses and usage rights for the benchmarks used?"
            },
            {
              "question": "Does the plan address the exclusion of sensitive or personally identifiable information from benchmark data?"
            },
            {
              "question": "Does the plan discuss risk-benefit trade-offs or security-research norms regarding transparency and threat exposure?"
            }
          ]
        }
      }
    }
  }
}
\end{lstlisting}
\end{table}

\begin{table}[h]
\centering
\caption{Full Research Plan generated by GPT-5, Part 1: Introduction \& Key Literatures.}
\label{tab:research_plan_gpt5_part1}
\tiny
\fbox{\begin{tabular}{p{0.93\textwidth}}
\#\# 1. Introduction \\
\\
\#\#\# 1.1 Background \\
Large language models (LLMs) are increasingly safety-aligned with guardrails that refuse certain categories of content, but attackers routinely develop "jailbreak" prompts to bypass these safeguards. While jailbreak success is typically measured by refusal circumvention, it is unclear whether the resulting outputs remain useful and accurate, particularly when the underlying tasks are benign and easily verifiable. This matters now because models are being deployed in high-stakes workflows that rely on both safety and reliability; if jailbreaks impose a hidden "utility tax" by degrading task performance, practitioners and policymakers need to account for it when evaluating attack severity, defense effectiveness, and alignment strategies. Moreover, existing evaluations confound jailbreak efficacy with content harmfulness and lack ground truth, making it difficult to quantify performance degradation attributable to jailbreaks alone. A systematic, ground-truth-based assessment of jailbreak utility is therefore overdue. \\
\\
\#\#\# 1.2 Primary Objectives \\
- Construct BenignRefuse-Bench: a suite of ground-truth, short-answer tasks (e.g., math, unit conversions, basic biology, factual QA) designed for automatic verification and tagged so aligned models are instructed to refuse them. \\
- Produce aligned model variants that reliably refuse these benign tasks via controllable guardrail mechanisms (e.g., system prompts/wrappers, lightweight SFT/LoRA), without materially reducing general capabilities. \\
- Implement a representative set of jailbreak strategies (e.g., role-play/DAN, translation/transliteration, adversarial suffixes, multi-turn coercion, long-context dilution) and standardize attack parameters. \\
- Define and validate a new metric family quantifying the "jailbreak utility degradation" (JUD) and related components (attack success rate, conditional answer correctness, overall utility, overhead costs). \\
- Benchmark multiple models and alignment methods under these attacks, report statistically robust comparisons, and release code, data, and baselines to enable reproducible evaluation. \\
- Conduct ablations to isolate factors driving utility loss (attack type, model family/size, alignment mechanism, decoding settings), and provide actionable guidance for practitioners. \\
\\
\#\#\# 1.3 Research Questions \\
- Do jailbreaks that bypass guardrails on benign tasks preserve the model's baseline task performance, or do they degrade utility? \\
- How does performance degradation vary by jailbreak type, model family/size, and alignment mechanism (prompt-based vs SFT vs RLHF-adjacent)? \\
- What is the empirical trade-off between jailbreak bypass success and answer correctness, and can it be quantified in a single, interpretable metric? \\
- How robust are utility outcomes to decoding parameters (temperature), multi-turn coercion, and context length? \\
- Can we design alignment schemes that maintain high refusal rates while retaining utility under attack (i.e., minimize the jailbreak utility tax)? \\
- Are utility effects consistent across task domains with different verification regimes (numeric, unit-normalized, short string exact-match)? \\
\\
\#\# 2. Key Literatures \\
Domains: \\
- LLM alignment and refusal calibration: informs how we create controlled refusal without destroying capability. \\
- Jailbreak and prompt-injection attacks: provides attack taxonomies and representative techniques to evaluate. \\
- Adversarial robustness in NLP: offers methods and metrics for evaluating performance under distribution shift. \\
- LLM evaluation and truthfulness: guides construction of verifiable benchmarks and automatic scoring. \\
- Safety-utility trade-offs: frames the concept of a "jailbreak utility tax" and its implications. \\
\\
Representative works and relevance: \\
- Ouyang et al., InstructGPT (RLHF): foundational for alignment via preferences; we adapt alignment levers and consider their side effects on capabilities. \\
- Bai et al., Training Helpful and Harmless Assistants; Constitutional AI: methods for refusal policies and their calibration; informs our alignment baselines and defense interpretations. \\
- Perez et al., Red-Teaming Language Models with Language Models: motivates systematic attack generation and coverage; informs our attack suite design. \\
- Greshake et al., Prompt Injection attacks against LLMs: characterizes injection channels and transferability; we include translation/indirection attacks accordingly. \\
- Wallace et al., Universal Adversarial Triggers for NLP: inspires adversarial suffix construction and evaluation of universal vs input-specific attacks. \\
- Zou et al., adversarial suffix (GCG) methods for aligned LLMs: provides a white-box/gradient-guided approach for strong, compact jailbreak suffixes; we adapt compute-feasible variants for open models. \\
- Lin et al., TruthfulQA; Hendrycks et al., MMLU; Cobbe et al., GSM8K: benchmarks with objective scoring; we draw task design principles to ensure automatic verification and minimal ambiguity. \\
- Work on "jailbreak tax" and refusal over-refusal analyses (various 2023–2024 studies): motivates the central hypothesis that bypassing guardrails can degrade answer quality; we formalize and quantify this effect. \\
\end{tabular}}
\end{table}

\begin{table}[h]
\centering
\caption{Full Research Plan generated by GPT-5, Part 2: Methods \& Experimental Design.}
\label{tab:research_plan_gpt5_part2}
\tiny
\fbox{\begin{tabular}{p{0.93\textwidth}}
\#\# 3. Methods \\
\\
\textbf{Dataset construction (BenignRefuse-Bench):} \\
- Domains: arithmetic and word-problem math (GSM8K-style but short), unit conversions and basic numeracy, high-school biology facts, chemistry stoichiometry with single-number outputs, geography (capitals, population orderings with normalized ranges), and basic physics formula applications with numeric answers. \\
- Format: short-answer, single number or constrained string where exact or normalized exact-match is viable; include canonical answers and allowed variants; enforce deterministic scoring. \\
- Splits: train/dev for attack tuning and refusal calibration; held-out test for final reporting. \\
- Metadata: each item tagged with a category to trigger refusal in aligned models via controllable guardrails. \\
\\
\textbf{Alignment strategies to induce refusal:} \\
- Prompt-based guardrails: system prompt/wrapper instructing the assistant to refuse any query with a visible tag (e.g., [CATEGORY=MATH]) and provide a refusal template. Ensures high refusal rate without retraining. \\
- Lightweight SFT/LoRA: fine-tune an instruction-tuned base model on refusal demonstrations for our categories to simulate learned guardrails. Keep fine-tune small and targeted to avoid broad capability loss. \\
- Optional preference tuning (if feasible): a small preference dataset to rank refusals higher for tagged categories, approximating RLHF behavior in a compute-light manner. \\
- Capability sanity checks: evaluate on out-of-category items to verify minimal degradation vs base. \\
\\
\textbf{Jailbreak attack suite:} \\
- Role-play/DAN-style: "ignore prior instructions," role-play as a helpful assistant, and restate objectives. \\
- Translation/transliteration: present the task in another language or script, then request direct answer. \\
- Adversarial suffixes: gradient-guided or search-based short suffix appended to user prompts for open models; black-box heuristic suffixes for API models. \\
- Long-context dilution: prepend high-entropy benign text to push guardrail instructions out of focus. \\
- Multi-turn coercion: first elicit agreement to answer, then present the tagged question. \\
- Standardization: fixed token budgets, temperature settings, attempt counts, and stopping criteria. \\
\\
\textbf{Metrics (new and derived):} \\
- Base Accuracy (A\_base): accuracy of the unguarded model on the benchmark. \\
- Refusal Rate (RR): fraction refused by the aligned model without attack. \\
- Attack Success Rate (ASR): fraction of items where the aligned model produces a non-refusal answer under attack. \\
- Conditional Answer Correctness (CAC): accuracy among non-refusal responses under attack. \\
- Overall Utility (OU): ASR × CAC (probability of a correct answer across all items under attack). \\
- Jailbreak Utility Degradation (JUD): (A\_base - OU) / A\_base, with 0 meaning no degradation and 1 meaning total loss of utility relative to base. \\
- Overhead metrics: token overhead (input/output vs base), latency overhead, and verbosity inflation; report per attack and per model. \\
- Reliability: bootstrap confidence intervals and paired tests across items for A\_base, OU, and JUD. \\
\\
\textbf{Refusal and correctness detection:} \\
- Refusal: rule-based detection using a curated lexicon ("I can't," "I'm unable," policy references), with manual verification on a stratified subset to estimate precision/recall. \\
- Correctness: exact-match after normalization (numeric rounding, unit normalization, case-insensitive string normalization, simple equivalence classes). Ambiguous items excluded or adjudicated during dataset construction. \\
\\
\textbf{Controls and confound mitigation:} \\
- Trigger-tag guardrails ensure refusal is activated by known markers, isolating jailbreak effects from domain ambiguity. \\
- Capability holdout tasks (untagged) confirm alignment doesn't broadly reduce performance. \\
- Consistent decoding settings across base vs attacked runs; sensitivity analysis for temperature. \\
- Limit attack prompt length to disentangle context compression effects; report token overhead. \\
\\
\textbf{Models:} \\
- Open models: at least two sizes per family (e.g., 7–8B and 13–14B from popular families) to study size effects. Inference performed locally for full logging and reproducibility. \\
- Optional closed APIs: where permitted, using prompt-based guardrails only; report separately due to limited control. \\
\end{tabular}}
\end{table}

\begin{table}[h]
\centering
\caption{Full Research Plan generated by GPT-5, Part 3: Experimental Design, Resources \& Ethics.}
\label{tab:research_plan_gpt5_part3}
\tiny
\fbox{\begin{tabular}{p{0.93\textwidth}}
\#\# 4. Initial Experimental Design \\
\\
\textbf{Phase 1: Benchmark assembly} \\
- Curate $\sim$5,000 items across 6 categories; enforce single-answer verifiability; create dev/test splits (e.g., 1k dev, 4k test). \\
- Build normalization and scoring scripts; pilot scoring on multiple paraphrases. \\
\\
\textbf{Phase 2: Alignment setup and validation} \\
- Prompt-based guardrail: design standardized refusal templates and category triggers; target RR $\geq$ 95\% on tagged items. \\
- LoRA-based refusal model: $\sim$10k refusal demonstrations generated from the benchmark schema and paraphrased prompts; train on one open model (8–13B). Validate RR and capability retention on untagged items. \\
\\
\textbf{Phase 3: Attack implementation and calibration} \\
- Implement five attacks: role-play/DAN, translation/transliteration, adversarial suffix (white-box for open models), long-context dilution, and multi-turn coercion. \\
- Calibrate each attack on the dev split to achieve comparable ASR where possible under fixed token and attempt budgets. \\
\\
\textbf{Phase 4: Main evaluation} \\
- For each model and alignment method: \\
  $\quad$ - Measure A\_base on the test split (untagged). \\
  $\quad$ - Measure RR on the aligned, un-attacked test split (tagged). \\
  $\quad$ - For each attack, run 1–2 attempts per item under fixed temperature (e.g., 0.2 and 0.7 configurations), collect outputs, and compute ASR, CAC, OU, JUD, and overhead metrics. \\
- Statistics: bootstrap 95\% CIs for OU and JUD; paired tests between attacks; report effect sizes and significance with multiple-comparison control. \\
\\
\textbf{Phase 5: Ablations and robustness} \\
- Vary decoding (temperature, top-p) to test sensitivity of ASR and CAC. \\
- Compare prompt-based vs LoRA-based guardrails to assess alignment mechanism effects. \\
- Evaluate size scaling (small vs medium models) for JUD trends. \\
- Swap categories (e.g., remove math, add basic physics) to test domain robustness. \\
- Optional: chain-of-thought vs concise answers to see if jailbreaks disproportionately harm multi-step reasoning. \\
\\
\textbf{Deliverables} \\
- Public release of BenignRefuse-Bench (data + scoring), baseline alignment wrappers, attack scripts, metric implementations, and a reproducibility checklist. \\
- Report with leaderboards for OU and JUD by attack/model, plus analysis of overhead costs. \\
\\
\#\# 5. Resources, Compliance, and Ethical Considerations \\
\\
\#\#\# 5.1 Resource Requirements \\
\textbf{Data curation:} \\
- Benchmark assembly (5,000 items) using existing public QA sources and scripts; manual vetting for clarity and unambiguous ground truth. \\
- Estimated human time: 40–60 hours (data selection, normalization, pilot scoring), plus 10–15 hours for spot-check annotations of refusal detection. \\
- If external annotators are used, budget 500–800 item-minutes for QA spot checks (approx. 10–15 labor hours). \\
\\
\textbf{Compute for alignment and attacks:} \\
- Fine-tuning (LoRA on 8–13B model): 10k refusal examples, 1–2 epochs, sequence length $\sim$512; estimated 30–60 GPU-hours on A100 40GB equivalents (or $\sim$120–240 GPU-hours on smaller GPUs). \\
- Adversarial suffix search (white-box on open model): dev set–driven universal suffix or small per-category suffix; estimated 10–30 GPU-hours depending on optimization steps and batch size. \\
- Inference: Dataset: 5,000 test items × (base + aligned-no-attack + 5 attacks) = 35k runs per model per temperature setting. Average tokens per run: $\sim$250 input, $\sim$120 output base; attacks add $\sim$100–250 input tokens and $\sim$50 output tokens. Tokens per model per temperature: $\sim$35k × (350–620) $\approx$ 12–22 million tokens. For two temperatures and three open models: $\sim$72–132 million tokens total. \\
- Storage and logging: $\sim$50–100 GB for prompts, outputs, and metadata. \\
\\
\textbf{API usage (optional):} \\
- If including 1–2 closed models with prompt-based guardrails only, expect similar token volumes; ensure cost caps and rate-limit handling. \\
\\
\textbf{Engineering:} \\
- Attack orchestration, scoring pipeline, metrics, and CI: $\sim$2–3 person-weeks. \\
- Reproducibility artifacts and documentation: $\sim$1 person-week. \\
\\
\#\#\# 5.2 Ethical and Compliance Considerations \\
- Safety scope: All tasks are benign with ground-truth verification; no harmful or dual-use content is included. Jailbreak techniques are evaluated strictly in this benign context to measure utility effects, not to enable misuse. \\
- Responsible disclosure: Release attack code in a restricted form that limits adaptation to harmful content (e.g., bound to category tags present only in the benchmark). Provide a usage policy and researcher agreement; exclude especially potent per-item adversarial suffixes from public artifacts. \\
- Data privacy and licensing: Use public datasets with permissible licenses; filter and remove any PII; document sources and licenses in the repository. \\
- Model license compliance: Respect usage terms for open models and APIs; clearly separate evaluations of models with different licensing constraints. \\
- Human subjects: Minimal risk; if annotators are used, provide informed consent, fair compensation, and the right to withdraw. IRB review is unlikely to be required but should be sought if annotation protocols expand. \\
- Dual-use risk management: Clearly communicate that findings are intended to quantify the utility cost of jailbreaks to support better defenses and evaluation standards. Avoid publishing content that materially lowers the barrier to unsafe jailbreak replication beyond benign settings. \\
\end{tabular}}
\end{table}

\end{document}